\documentclass[11pt]{article}
\usepackage{amsmath}
\usepackage{booktabs}
\usepackage[final]{acl}
\usepackage{times}
\usepackage{latexsym}
\usepackage[T1]{fontenc}
\usepackage[utf8]{inputenc}
\usepackage{microtype}
\usepackage{inconsolata}
\usepackage{graphicx}
\usepackage{multirow}
\usepackage{xcolor}

\title{Beyond Prompts: Measuring and Optimizing LLM Tool-Agent Harnesses}
\author{
  \textbf{Cen Mia Zhao} \quad \textbf{Haibo Ruan} \quad \textbf{Wenjie Chen} \\
  \textbf{Pei-fen Tu} \quad \textbf{Usman Abbasi} \quad \textbf{Joel Hesch} \\
  Airbnb, Inc., USA \\
  \texttt{\{mia.zhao, haibo.ruan, wenjie.chen, pei-fen.tu,} \\
  \texttt{usman.abbasi, joel.hesch\}@airbnb.com}
}
\begin{document}
\maketitle
\begin{abstract}
LLM tool agents can be improved without retraining by modifying the runtime
harness around a fixed model: prompts, tool interfaces, middleware, state
handling, and recovery logic. We study this setting as resource-bounded
harness selection for fixed-model multi-turn tool agents, with the search
surface scoped to prompts and \emph{tool-boundary} middleware: edits are
guarded intercepts at the tool boundary, not arbitrary rewriting of agent
execution logic. Our optimizer-agnostic protocol reports mean held-out lift,
worst-condition lift, repeatability, logged cost diagnostics, and
\(\mathrm{RelLift}_{95}(B)\), a conservative estimate of the held-out gain
of the harness selected under budget \(B\). We instantiate the protocol with
prompt-only and prompt-plus-middleware optimizers, including \textsc{PRISM},
which clusters failures and routes repairs to prompt, tool-boundary
middleware, or joint edit surfaces within a Pareto search. On BFCL
multi-round, \(\tau^2\)-Retail, and \(\tau^2\)-Telecom, \textsc{PRISM}
obtains mean held-out lifts of 14.2, 14.9, and 10.1 percentage points and
positive empirical \(\mathrm{RelLift}_{95}\) on all three benchmarks, and a
component ablation attributes the margin chiefly to failure-surface routing
and the edit-pattern constraint. Across optimizers, the results show that
some search procedures can occasionally find large gains but still choose
brittle updates, so the reliability of the chosen harness should be reported
alongside average held-out lift.
\end{abstract}
\section{Introduction}
\label{sec:introduction}
Multi-turn conversational agents that use external tools are a demanding test
bed for agent optimization: in customer-facing service workflows, agents must
track dialogue state, follow policies, call APIs correctly, recover from
missing information, and leave the environment in the intended
state. \(\tau^2\)-bench and the Berkeley Function
Calling Leaderboard (BFCL) capture complementary aspects of this setting
(dynamic user interaction, stateful execution, executable function
calling)~\citep{barres2025tau2benchevaluatingconversationalagents}
and expose a deployment
tension: stronger systems can be more reliable, but production agents must
still satisfy latency and runtime-cost
constraints~\citep{chen2024frugalgpt}.
We study whether \emph{agent harness optimization} can improve this trade-off
without changing model weights. A harness is the runtime control layer around a
fixed target model, including prompts, tool interfaces, context construction,
state management, middleware, recovery logic, and evaluation hooks. Prior work
on instruction optimization, trajectory-reflective prompt evolution, and
reasoning-and-acting agents shows that changing this surrounding layer can
substantially affect downstream behavior
~\citep{opsahlong2024mipro,agrawal2026gepa,yao2023react}. We scope the
optimized surface precisely to prompts plus \emph{tool-boundary} middleware:
middleware helps when failures are locally observable and checkable at the
tool boundary, not via arbitrary rewriting of agent execution logic.
The main question is not whether one harness can improve one benchmark score,
but whether an automatic harness search procedure can be trusted: a reported
gain may depend on the repair examples used during search, stochastic optimizer
choices, or variance in agentic evaluation, and may carry search cost that
limits practical value~\citep{bjarnason2026randomness}. We therefore evaluate
harness search procedures along four axes:
\textbf{effectiveness}, held-out lift over a baseline harness;
\textbf{repair-distribution robustness}, whether lift persists across repair
sets; \textbf{same-protocol repeatability}, whether the effect recurs under the
same search protocol; and \textbf{efficiency}, the budget required to obtain the
gain reliably. We instantiate this protocol on three multi-turn tool-use tasks:
BFCL multi-round, \(\tau^2\)-Retail, and \(\tau^2\)-Telecom \citep{barres2025tau2benchevaluatingconversationalagents} with
user simulation. We compare BetterHarness-style hill climbing, MIPROv2 \citep{opsahlong2024mipro}, GEPA \citep{agrawal2026gepa},
and PRISM, our factorized optimizer over prompt and middleware updates.
This paper makes three contributions. First, we frame harness improvement as
a budgeted selection problem and ask how reliable the selected harness is on
held-out data, measured with the \(\mathrm{RelLift}_{95}\) estimator.
Second, PRISM routes each repair to the surface where it can be locally
checked, based on clustering failures by root cause. Third, PRISM selects
candidates in the loop by both gate pass rate and reliability, not score
alone. The evolutionary search, Pareto retention, and failure clustering
themselves are not new: PRISM builds them on a GEPA-style skeleton
\citep{agrawal2026gepa} and runs on the same execution substrate as every
baseline. The component ablation (\S\ref{sec:ablation-component}) checks
whether the new pieces, not the inherited ones, carry the result.
\section{Background and Related Work}
\label{sec:related}
Prior work shows that the harness around a model is an important optimization
surface: BetterHarness treats evaluation failures as training data for harness
hill climbing, with holdout evaluation and regression checks before
deployment~\citep{trivedy2026betterharness}, and Deep Agents profiles expose
the surface explicitly, with profile-specific gains on \(\tau^2\)-bench-style
tasks~\citep{deepagents2025,langchain2026benchmark}.
One line of work optimizes the text around a fixed model: OPRO uses the model
itself as an instruction optimizer~\citep{yang2024opro}, EvoPrompt applies
evolutionary operators to prompt populations~\citep{guo2024evoprompt},
TextGrad propagates textual feedback through compound
systems~\citep{yuksekgonul2024textgrad}, and DSPy compiles language-model
programs with MIPROv2 as its instruction
optimizer~\citep{khattab2024dspy,opsahlong2024mipro}. GEPA evolves prompts by
trajectory reflection over a Pareto pool and reports outperforming these
earlier methods~\citep{agrawal2026gepa}, so we baseline against GEPA and
MIPROv2 rather than every ancestor.
A second line edits agent \emph{code}: ADAS searches over agent
programs~\citep{hu2024adas}, Meta-Harness rewrites harness
source~\citep{lee2026metaharness}, and Trace optimizes the agent computation
graph~\citep{cheng2024trace}. These systems work offline on coding, math, or
classification tasks, search an unconstrained program space, and place no
resource limits in the objective. Our setting differs on all three axes:
multi-turn conversational tool agents with a simulated live user, constrained
tool-boundary repair around a fixed model and tool API, and a deployment
envelope inside the score itself (Section~\ref{sec:budgets}). NeoSigma
studies automatic harness repair in a multi-turn
setting~\citep{gupta2026selfimproving}.
These works establish that harness search can improve agent behavior, but they
report different pieces of evidence: validation lift, rollout efficiency,
regression checks, or a single final score. In repeated search-and-release,
reporting a single validation lift is not enough: a team maintaining a
deployed agent receives failure reports in batches, and this week's batch
differs from last week's in distribution and often in
kind~\citep{gama2014drift}. Such a team needs a process that
is reliable across batches; we therefore evaluate the search procedure
itself, along the four axes of Section~\ref{sec:introduction}.
\section{Evaluation Protocol and Metrics}
\label{sec:metrics}
We evaluate optimizer procedures that return harnesses for a fixed target model.
The protocol is optimizer-agnostic: it treats each procedure as a black box that
returns a candidate harness, declares its pre-scorecard search cost, and is then
evaluated on held-out scorecard data.
\paragraph{Split roles.}
Each run partitions benchmark data into three roles, used throughout the
paper. The \emph{repair} split \(D_{\mathrm{repair}}\) is used to propose and
test harness edits. The \emph{gate} split \(D_{\mathrm{gate}}\) is used to
accept candidates and to make the final pre-scorecard selection. The
\emph{scorecard} split \(D_{\mathrm{score}}\) is held out from the entire
search process and touched exactly once, for final reporting. Development data
means the union of repair and gate data. Unless otherwise stated, the baseline
and returned harness are evaluated on the same scorecard cases, so lift is a
paired comparison.
\paragraph{Benchmark interface.}
For benchmark \(b\), harness \(h\), data split \(D\), and reported metric
\(m\), let \(S_{b,m}(h,D)\in[0,1]\) be the benchmark-native score, averaged
over per-instance scores when the benchmark reports them. Invalid executions
(crashes, timeouts, invalid actions, protocol violations) are counted
according to the benchmark's failure rule.
\paragraph{Effectiveness.}
Let \(h_0\) be the baseline harness. For run \(i\), let \(h_i\) be the harness
returned before scorecard evaluation. The held-out scorecard lift is
\[
\Delta_{i,m}
=
S_{b,m}(h_i,D_{\mathrm{score},i})
-
S_{b,m}(h_0,D_{\mathrm{score},i}),
\]
reported in percentage points. \(\mathrm{MeanLift}_m\) is the average paired
lift over runs, reported with paired confidence intervals.
\paragraph{Stability and repeatability.}
To measure whether improvement depends on the development data, we repeat the
same protocol over pre-specified conditions: a random split, a failure
source, a domain, or a time batch of arriving production failures.
\(\mathrm{WorstLift}_m\) is the lowest mean lift across conditions (the
weakest batch), and \(\mathrm{RepRate}_{\delta,m}\) is the fraction of runs
with lift at least \(\delta\), with \(\mathrm{RepRate}_{0,m}\)
(\(\mathrm{RR}_0\)) the fraction of runs with positive held-out lift.
\paragraph{Efficiency and budgeted reliable lift.}
Efficiency asks how much held-out lift a procedure can reliably obtain under a
fixed pre-scorecard search budget. For run \(i\), let \(C_i\) be the
pre-scorecard search cost under a declared accounting convention, covering
search-time activity (proposal, candidate evaluation, filtering, repair,
and selection) before final scorecard evaluation; the final held-out
scorecard evaluation is always excluded, and the components a given
convention prices are stated with it (ours are defined in
Appendix~\ref{app:cost-methodology}). Let \(\bar C\) be the mean cost over runs.
Under budget \(B\), the number of affordable independent optimizer runs is
\(N_B=\lfloor B/\bar C\rfloor\). Each run returns a harness \(h_i\) and a
pre-scorecard selection statistic \(G_i\) (gate score, validation score, or
the optimizer's internal pre-scorecard criterion) computed without access to
\(D_{\mathrm{score}}\). Under budget \(B\) we select
\(i^*(B)=\arg\max_{i\in\mathcal R_B} G_i\) over the \(N_B\) affordable runs,
and the held-out lift of the selected harness is
\(L_B=\Delta_{i^*(B),m}\). Our budgeted reliable-lift metric is
\[
\mathrm{RelLift}_{\gamma}(B)
=
Q_{1-\gamma}(L_B),
\]
the lower \((1-\gamma)\)-quantile of selected held-out lift over repeated uses
of the same protocol, estimated by bootstrap over observed optimizer runs:
each bootstrap sample draws \(N_B\) runs with replacement, selects the run
with the highest pre-scorecard statistic \(G_i\), and records the selected
run's held-out lift. \(\mathrm{RelLift}\) is a \emph{deployer-side}
diagnostic, deliberately outside the optimizer loop: an empirical lower-tail
estimate, not a coverage guarantee over future scorecards, simulator seeds,
or repair distributions. Results selected using scorecard information are
reported only as oracle ceilings.
\section{Experiments}
\label{sec:experiments}
Each optimizer searches under a fixed rollout budget, selects one harness
using only pre-scorecard information, and is evaluated once on held-out
scorecard cases. The main comparison uses \texttt{gpt-5.4-mini} as the inner
execution model and \texttt{claude-opus-4.7} as the outer proposal model;
Section~\ref{sec:ablation_inner_model} varies each role with the other
fixed.
\subsection{Benchmarks}
\label{sec:benchmarks}
We evaluate on BFCL multi-round and two domains from \(\tau^2\)-bench:
\(\tau^2\)-Retail and \(\tau^2\)-Telecom. BFCL tests executable multi-step
function calling: a case passes if the final tool-call state matches the
benchmark oracle. The \(\tau^2\)-bench domains test simulated service agents
with dialogue state, task policies, and tool-mediated state changes; a case
passes if the final simulator outcome satisfies the benchmark success criterion.
We report BFCL with \(\mathrm{pass}^{1}\) and the \(\tau^2\) domains with the
benchmark's \(\mathrm{pass}^{4}\) implementation (four simulator rollouts per
task). The tasks stress different failure modes: BFCL emphasizes
schema-correct calling under a fixed API, while the \(\tau^2\) domains add
longer-horizon policy following, state tracking, missing-information
recovery, and simulator stochasticity.
\subsection{Run protocol}
\label{sec:run-protocol}
Each optimizer run uses disjoint repair, gate, and scorecard splits with the
roles defined in Section~\ref{sec:metrics}.
For BFCL, each run uses 100 repair cases, 100 gate cases, and a
disjoint 600-case scorecard split, with splits independently re-drawn per
seed. For each \(\tau^2\)-bench domain, the scorecard split contains 74 task
instances; repair and gate sizes are reported in
Appendix~\ref{app:run-config}.
On the \(\tau^2\) domains, the four
split seeds are drawn so that the four repair sets are pairwise disjoint and
the four gate sets are pairwise disjoint (\(4\times 20=80\) of each
domain's 114 tasks per role), so every run must improve the harness from a
different batch of failure cases, simulating the disjoint error batches a
deployed team receives over time. Within a seed, repair, gate, and scorecard
splits are disjoint; across seeds, because each seed's scorecard reserves 74
of the 114 tasks, one seed's repair set may overlap another seed's gate or
scorecard set. Under this reading, \(\mathrm{WorstLift}\)
is the weakest batch and
\(\mathrm{RR}_0\) the fraction of batches the procedure still delivered on.
Each benchmark--optimizer pair targets
four split seeds and up to four independent repeats per seed under the same
rollout budget, a design maximum of 16 runs. Runs that fail the reliability
screen of Appendix~\ref{app:run-config} are excluded as infrastructure
failures, so per-arm denominators \(n\le 16\) vary by arm; all reported
statistics are computed over an arm's surviving \(n\) runs unless otherwise
stated.
\paragraph{Baseline harness.}
The baseline \(h_0\) is the \emph{native Deep Agents setup}: every agent is
instantiated with \texttt{create\_deep\_agent} from the \texttt{deepagents}
library (v0.5.6; Appendix~\ref{app:compute-setup}), unmodified, with
benchmark-native tools and the benchmark's default system prompt. Every lift
in every table is therefore measured against native Deep Agents on these
exact benchmarks. On the scorecard splits, \(h_0\) attains
51.0 \(\mathrm{pass}^{1}\) on BFCL;
18.2 \(\mathrm{pass}^{4}\) (47.6 \(\mathrm{pass}^{1}\)) on
\(\tau^2\)-Retail; and
6.1 \(\mathrm{pass}^{4}\) (25.0 \(\mathrm{pass}^{1}\)) on
\(\tau^2\)-Telecom, pooled over the four per-seed scorecard splits. The gap
between \(\mathrm{pass}^{1}\) and \(\mathrm{pass}^{4}\) reflects the
strictness of the all-trials success rule under simulator stochasticity.
All absolute scores in this paper, including the \(h_0\) baselines above,
are computed on the held-out scorecard splits (per seed, 74 of each
\(\tau^2\) domain's 114 tasks and a 600-case BFCL split) rather than on
the full public task sets, so they are not directly comparable to numbers
reported on the complete benchmarks.
For orientation, \citet{barres2025tau2benchevaluatingconversationalagents}
report Telecom \(\mathrm{pass}^{1}\) of 34--49\% for 2025 frontier models
under the benchmark's native agent loop; these figures are not directly
comparable to ours, which use a smaller inner model, an unmodified generic
harness, and per-seed scorecard splits, and our Telecom
\(\mathrm{pass}^{1}\) baseline falls below that range, as expected for a
smaller inner model under a generic harness.
\paragraph{Fairness of gate-split usage.}
All optimizers receive identical splits and draw gate rollouts from one fixed
inner-rollout budget (Section~\ref{sec:budgets}); none sees the scorecard
before final evaluation, and they differ only in \emph{when} they consult
the gate (Table~\ref{tab:optimizer_comparison}): BH at every iteration,
GEPA and MIPROv2 once, at final acceptance, matching their published forms
\citep{agrawal2026gepa,opsahlong2024mipro}. The selection statistic \(G_i\)
(Section~\ref{sec:budgeted-estimator}) is computed identically for every
optimizer's returned harness, and no baseline is denied data its published
form uses; per-optimizer split usage and exact wrapper configurations are
in Appendix~\ref{app:wrapper-config}.
\subsection{Optimizers}
\label{sec:optimizers}
\begin{figure*}[t]
    \centering
    \includegraphics[width=0.68\textwidth]{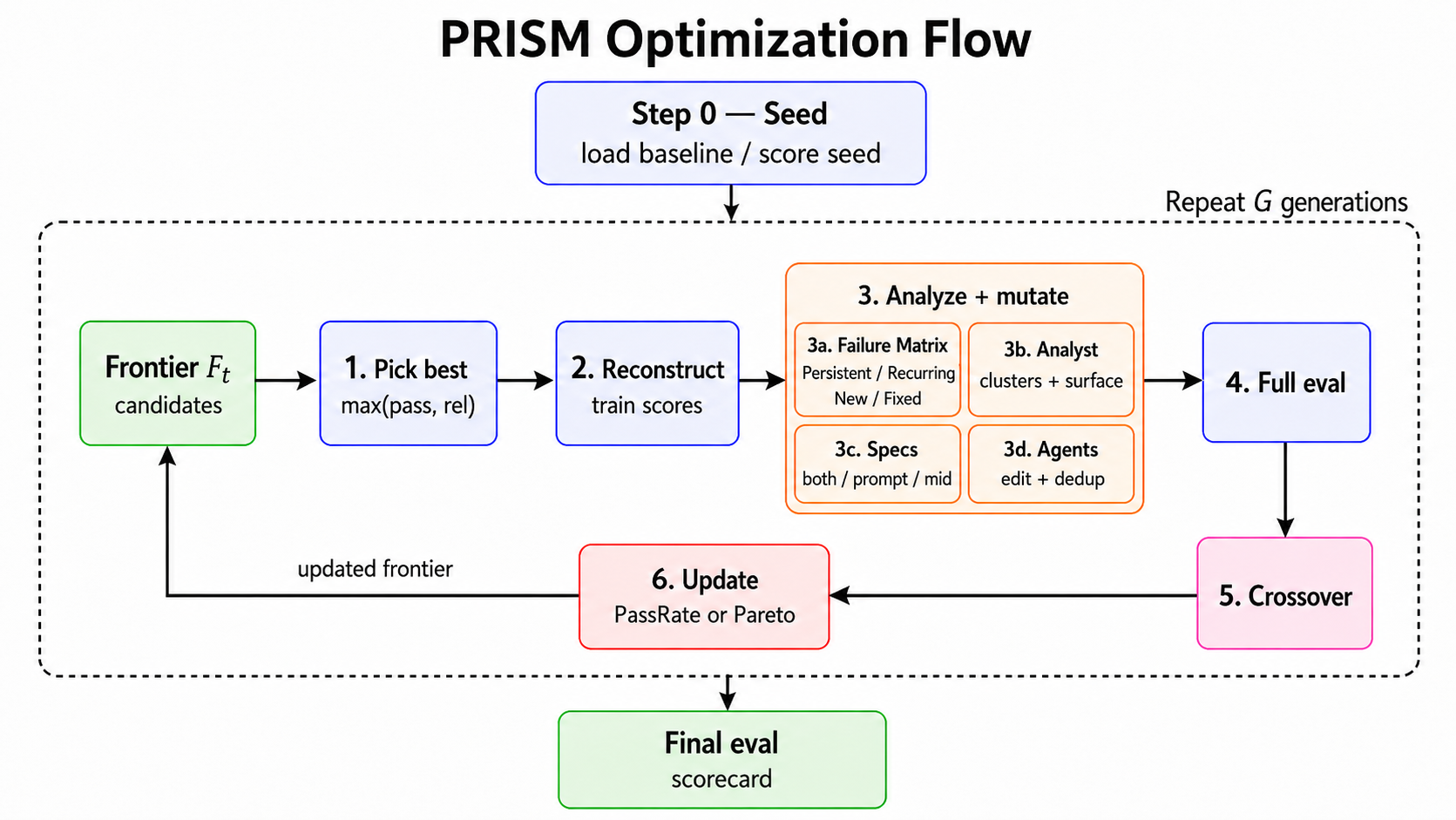}
    \caption{
    The PRISM loop (\S\ref{sec:optimizers}). Each generation, an analyst
    LLM clusters failing cases by root cause and routes each cluster to the
    surface where it can be fixed: prompt, middleware, or both. Candidate
    harnesses evolve through mutation and crossover, a Pareto frontier
    retains the best trade-offs between gate pass rate and reliability, and
    the selected harness is evaluated once on the held-out scorecard.
    }
    \label{fig:prism-flow}
\end{figure*}

We compare optimizers by editable surface and selection state. Prompt-only
methods edit the system-prompt surface; prompt-plus-middleware methods may
also edit execution-time middleware components at the tool boundary.
\emph{Naming:} \textsc{X-PO} / \textsc{X-MW} denote each
optimizer's prompt-only / prompt+middleware variant;
\textsuperscript{\dag}~marks each baseline's as-published form
(\textsc{BH-MW}\textsuperscript{\dag}, \textsc{GEPA-PO}\textsuperscript{\dag},
\textsc{MIPROv2-PO}\textsuperscript{\dag}), and unmarked variants are
surface-toggled counterparts we construct symmetrically. Where unambiguous,
we write \textsc{PRISM} for \textsc{PRISM-MW} and \textsc{BH} for
\textsc{BH-MW}.
Table~\ref{tab:optimizer_comparison} summarizes the procedures;
Figure~\ref{fig:prism-flow} illustrates PRISM. Each baseline runs in its published form: we defer to the original
papers for full algorithms and document our exact wrapper configurations in
Appendix~\ref{app:wrapper-config}.
\paragraph{Middleware edit patterns.}
Across all prompt-plus-middleware optimizers except BH (kept unconstrained
for fidelity to its as-published form), middleware mutations are constrained
to three tool-boundary edit patterns: \emph{silent correction}, rewriting
malformed arguments before execution; \emph{error blocking}, returning an
explicit tool error so the model can retry; and \emph{prerequisite blocking},
checking conversation or state history before allowing a call. The patterns
fix \emph{where} an edit acts and \emph{what shape} it takes, but the
predicate and repair logic inside each guard are open-ended programs
generated per failure cluster; repairs requiring cross-call coordination,
retries, asynchronous state, or control-flow changes fall outside this
vocabulary (see Limitations). GEPA-MW and MIPROv2-MW receive these patterns
exactly as PRISM's middleware slot does, so comparisons isolate search
machinery rather than edit vocabulary.
\paragraph{PRISM.}
PRISM is a failure-routed evolutionary optimizer over a bounded Pareto
frontier (population cap five, ten generations by default). Each generation
(1)~\emph{selects} the best viable frontier parent; (2)~\emph{analyzes}
failures with an analyst LLM that groups failing cases by root cause from
tool sequences, state diffs, and error traces, labeling each cluster's fix
surface as \textsc{Prompt}, \textsc{Middleware}, or \textsc{Joint};
(3)~\emph{routes} clusters to three surface-constrained mutation slots: a
joint slot that may edit both surfaces, a prompt slot restricted to one or
two universal behavioral rules, and a middleware slot restricted to the three
edit patterns above; (4)~\emph{evaluates} children on the repair and gate
splits, running \emph{crossover} when one child solves a case the best child
still fails; and (5)~\emph{updates} a Pareto frontier over gate pass rate
and reliability \(R=1-\mathrm{Stuck}\), where a candidate dominates only if
it is no worse on both axes and strictly better on one. A cross-generation
\emph{failure matrix} labels each case \textsc{New}, \textsc{Recurring},
\textsc{Persistent}, or \textsc{Fixed}; mutators prioritize recurring cases,
protect fixed cases, and never repeat a rejected approach without a new
angle. PRISM-PO forces all slots to prompt-only. This subsection is the complete
algorithm; Appendix~\ref{app:prism-process} adds the analyst prompt, the
slot access rules, the mutator workspace and safety checks, and frontier
tie-breaking.
\begin{table*}[t]
\centering
\footnotesize
\setlength{\tabcolsep}{3.5pt}
\renewcommand{\arraystretch}{1.05}
\caption{How the four optimizers search.
\textsuperscript{\dag}~marks an optimizer whose published form we run
unchanged. The last column describes how each \textsc{-MW} variant may edit
middleware; the \textsc{-PO} variants edit prompts only. Terms are defined
in \S\ref{sec:run-protocol}--\S\ref{sec:optimizers}; all optimizers share
the budget of \S\ref{sec:budgets}.}
\label{tab:optimizer_comparison}
\begin{tabular}{@{}p{0.08\textwidth}p{0.17\textwidth}p{0.06\textwidth}p{0.16\textwidth}p{0.25\textwidth}p{0.14\textwidth}@{}}
\toprule
Optimizer & Search & Failure matrix & Clustering \(\to\) routing & Gate use in loop & Middleware edits \\
\midrule
\textsc{BH}\textsuperscript{\dag}      & Greedy hill climbing & Yes & None & Each iteration & Unconstrained \\
\textsc{GEPA}\textsuperscript{\dag}    & Reflective evolution, Pareto pool & No & None & Final acceptance only & Pattern-guided \\
\textsc{MIPROv2}\textsuperscript{\dag} & Bayesian search & No & None & Final acceptance only & Pattern-guided \\
\textsc{PRISM}   & Evolution + crossover, Pareto frontier & Yes & Clusters routed to surface slots & Per generation, by (gate rate, reliability) & Pattern-guided, routed \\
\bottomrule
\end{tabular}
\end{table*}
This design separates search strategy from search surface: \textsc{BH-PO}
and \textsc{PRISM-PO} isolate the effect of exposing middleware to
optimization, and \textsc{GEPA-MW} and \textsc{MIPROv2-MW} isolate PRISM's
search machinery at an identical middleware surface.
\subsection{Budgets, resources, and artifacts}
\label{sec:budgets}
All search-budget quantities exclude final scorecard evaluation. The
controlled resource is the number of inner-model task rollouts during search,
identical across optimizers. Monetary Cost/run is a
logged implementation diagnostic.
Each task rollout must also finish within a benchmark-specific resource
envelope of wall-clock, maximum-turn, and token caps
(Appendix~\ref{app:run-config}); rollouts that exceed the envelope, crash,
enter repeated tool-call loops, or terminate abnormally
receive zero score, reflecting the deployment goal that the agent must finish
within allocated resources.
Our code is publicly available, containing PRISM, baseline optimizer wrappers,
evaluation runners, and metric aggregation
scripts.\footnote{\url{https://github.com/airbnb/agent-harness-optimizer}}
Appendix~\ref{app:reproducibility} gives additional implementation details.
\subsection{Budgeted reliable-lift estimator}
\label{sec:budgeted-estimator}
For each benchmark--optimizer pair, we observe \(n\le 16\) run tuples
\(\{(C_i,G_i,\Delta_i)\}_{i=1}^{n}\) and apply the
\(\mathrm{RelLift}_{95}(B)\) estimator of Section~\ref{sec:metrics} with
5{,}000 bootstrap resamples; the resamples reduce Monte Carlo error and do not
increase the number of independent optimizer runs beyond the \(n\) observed. In
our implementation, the pre-scorecard statistic \(G_i\), computed
identically for every optimizer's returned harness, orders candidates by
gate-split pass rate, breaking ties by reliability:
\[
G_i=\big(S_{\mathrm{gate}}(h_i,D_{\mathrm{gate}}),
1-\mathrm{Stuck}(h_i,D_{\mathrm{gate}})\big).
\]
A rollout counts as stuck if it violates the resource envelope of
\S\ref{sec:budgets}; remaining ties break deterministically by run
identifier.
The subscript in \(\mathrm{RelLift}_{95}\) is a quantile level, not a
confidence level: it identifies the 5th percentile of selected held-out
lift, and says nothing by itself about how precisely a small run pool pins
that percentile down. Because the estimate rests on at most 16 underlying
runs, we quantify its uncertainty for the \textsc{PRISM-MW} arms in two
ways: a percentile-bootstrap confidence interval on each
\textsc{PRISM-MW} \(\mathrm{RelLift}_{95}\) value (obtained by resampling
that arm's run tuples and recomputing the full estimator), and a
subsampling sensitivity analysis; both are reported in
Appendix~\ref{app:subsampling}
(Table~\ref{tab:rellift-uncertainty}), and the released estimator computes
the same quantities for any arm from its run tuples. Baseline
\(\mathrm{RelLift}_{95}\) values are reported as point estimates, so
cross-optimizer \(\mathrm{RelLift}_{95}\) comparisons rest on point
estimates and are always read alongside \(\mathrm{WorstLift}\) and
\(\mathrm{RR}_0\), which estimate the same tail from different statistics;
where intervals are available, we do not describe
\(\mathrm{RelLift}_{95}\) differences smaller than the estimator's own
interval as differences.
\begin{table*}[t]
\centering
\scriptsize
\setlength{\tabcolsep}{4.0pt}
\renewcommand{\arraystretch}{1.0}
\caption{
Main held-out results. Within each benchmark, prompt-only
(\textsc{-PO}) and prompt+middleware (\textsc{-MW}) variants are grouped
separately, and bold marks the best value in that group (for cost, the
lowest). All lifts are percentage points over the native Deep Agents
baseline \(h_0\) (\S\ref{sec:run-protocol}), measured by
\(\mathrm{pass}^{1}\) on BFCL and \(\mathrm{pass}^{4}\) on the \(\tau^2\)
domains. \(\mathrm{RR}_{0}\) is the fraction of runs with positive lift.
\(\mathrm{RelLift}_{95}\) answers: if a deployer could afford several runs
and picked one using only pre-scorecard information, how much lift would
the picked harness deliver in the unlucky (5th-percentile) case
(\S\ref{sec:metrics}, at budget \(B_{\mathrm{ref}}\)). Cost/run (inner)
prices inner-model rollout calls only
(Appendix~\ref{app:cost-methodology}). Per-arm run counts \(n\) can fall
below 16 under the exclusion rule (Appendix~\ref{app:run-config}).
\textsuperscript{\dag}~marks an optimizer run in its published form.
\textsuperscript{\S}~marks the two Telecom rows scored from four simulator
trials (one original + three run afterwards) combined outside the
benchmark-native \(\mathrm{pass}^{4}\) path; other Telecom rows use the
native path (see Limitations).
}
\label{tab:main-scorecard-results}
\resizebox{\textwidth}{!}{
\begin{tabular}{
@{} p{0.12\textwidth} l r l r r r r r @{}
}
\toprule
Benchmark & Optimizer & MeanLift & 95\% CI & MaxLift & WorstLift
& \(\mathrm{RR}_{0}\) & \(\mathrm{RelLift}_{95}\) & Cost/run (inner) \\
\midrule
\multirow{10}{*}{\shortstack[l]{BFCL\\600-case\\\(\mathrm{pass}^{1}\)}}
& \multicolumn{8}{@{}l}{\emph{Prompt-only}} \\
& BH-PO       & \textbf{7.4} & [6.1, 8.8]   & \textbf{9.3} & \textbf{5.7} & \textbf{100\%} & \textbf{6.4} & \$127 \\
& GEPA-PO\textsuperscript{\dag}    & 2.0 & [1.7, 2.2]   & 2.3          & 1.5          & \textbf{100\%} & 1.7          & \textbf{\$15} \\
& MIPROv2-PO\textsuperscript{\dag} & 5.8 & [3.6, 8.3]   & \textbf{9.3} & 3.0          & \textbf{100\%} & 4.1          & \$49 \\
& PRISM-PO    & 5.8          & [3.9, 7.7]   & 8.7          & 3.0          & \textbf{100\%} & 3.9          & \$79 \\
\cmidrule(l){2-9}
& \multicolumn{8}{@{}l}{\emph{Prompt + middleware}} \\
& BH-MW\textsuperscript{\dag} & 10.8 & [8.4, 13.2]  & 14.0          & 7.5           & \textbf{100\%} & 9.1           & \$167 \\
& GEPA-MW     & 5.3           & [2.3, 8.2]   & 11.3          & 1.7           & \textbf{100\%} & 3.7           & \textbf{\$27} \\
& MIPROv2-MW  & 7.6           & [4.1, 11.1]  & 12.3          & 3.8           & \textbf{100\%} & 6.1           & \$76 \\
& PRISM-MW    & \textbf{14.2} & [11.7, 16.7] & \textbf{17.5} & \textbf{11.0} & \textbf{100\%} & \textbf{12.0} & \$103 \\
\midrule
\multirow{10}{*}{\shortstack[l]{\(\tau^2\)-Retail\\74-case\\\(\mathrm{pass}^{4}\)}}
& \multicolumn{8}{@{}l}{\emph{Prompt-only}} \\
& BH-PO       & 4.7           & [-1.2, 10.6] & \textbf{23.0} & -13.5        & 73\%           & -0.4         & \$96 \\
& GEPA-PO\textsuperscript{\dag}    & -4.7 & [-9.5, 0.0]  & 1.4           & -12.2        & 25\%           & -9.5         & \textbf{\$2} \\
& MIPROv2-PO\textsuperscript{\dag} & \textbf{6.1} & [3.0, 8.8] & 9.5     & \textbf{1.4} & \textbf{100\%} & \textbf{3.4} & \$22 \\
& PRISM-PO    & 1.8           & [-9.5, 12.2] & 12.2          & -9.5         & 67\%           & -5.4         & \$35 \\
\cmidrule(l){2-9}
& \multicolumn{8}{@{}l}{\emph{Prompt + middleware}} \\
& BH-MW\textsuperscript{\dag} & 3.2 & [-2.0, 8.9]  & \textbf{23.0} & -13.5        & 58\%           & -1.2          & \$112 \\
& GEPA-MW     & 1.4           & [-1.5, 4.3]  & 6.4           & -5.2         & 67\%           & 0.7           & \textbf{\$11} \\
& MIPROv2-MW  & 8.3           & [3.5, 13.2]  & 19.5          & 2.6          & \textbf{100\%} & 7.1           & \$37 \\
& PRISM-MW    & \textbf{14.9} & [8.4, 19.9]  & 21.6          & \textbf{5.4} & \textbf{100\%} & \textbf{10.1} & \$13 \\
\midrule
\multirow{10}{*}{\shortstack[l]{\(\tau^2\)-Telecom\\74-case\\\(\mathrm{pass}^{4}\)}}
& \multicolumn{8}{@{}l}{\emph{Prompt-only}} \\
& BH-PO       & 6.1          & [0.7, 11.5]  & 12.2          & -2.7         & 75\%           & 0.7          & \$82 \\
& GEPA-PO\textsuperscript{\dag}    & 0.0 & [-7.4, 7.4]  & 10.8          & -8.1         & 50\%           & -7.1         & \textbf{\$6} \\
& MIPROv2-PO\textsuperscript{\dag} & 6.3 & [1.4, 14.9]  & 14.9          & \textbf{1.4} & \textbf{100\%} & 1.8          & \$24 \\
& PRISM-PO\textsuperscript{\S}    & \textbf{6.8} & [1.4, 14.2]  & \textbf{17.6} & 0.0          & 75\%           & \textbf{2.0} & \$25 \\
\cmidrule(l){2-9}
& \multicolumn{8}{@{}l}{\emph{Prompt + middleware}} \\
& BH-MW\textsuperscript{\dag} & 0.3 & [-10.8, 14.5] & \textbf{20.3} & -13.5        & 25\%           & -8.1         & \$138 \\
& GEPA-MW     & 4.4           & [-2.4, 6.4]   & 11.8          & -4.1         & 50\%           & 0.7          & \textbf{\$6} \\
& MIPROv2-MW  & 7.3           & [1.7, 12.9]   & 13.3          & 0.7          & \textbf{100\%} & 3.3          & \$27 \\
& PRISM-MW\textsuperscript{\S}    & \textbf{10.1} & [7.4, 12.8]   & 13.5          & \textbf{6.8} & \textbf{100\%} & \textbf{7.8} & \$18 \\
\bottomrule
\end{tabular}
}
\end{table*}
\section{Results}
\label{sec:results}
Table~\ref{tab:main-scorecard-results} reports the main held-out results.
Among the prompt-only optimizers, no single strategy wins everywhere: BH-PO
is best on BFCL, MIPROv2-PO on Retail, and PRISM-PO on Telecom, while
GEPA-PO trails on all three tasks. Exposing the pattern-guided middleware
surface helps every optimizer that receives it, improving
\(\mathrm{RelLift}_{95}\) from the \textsc{-PO} to the \textsc{-MW}
variant in all nine GEPA, MIPROv2, and PRISM cells. The constraint matters
as much as the surface, however: BH, whose middleware edits are
unconstrained, falls below its own prompt-only variant on both \(\tau^2\)
tasks (\S\ref{sec:results_prompt_vs_middleware}). PRISM-MW ends up
strongest on every task, and the component ablation in
\S\ref{sec:ablation-component} traces this margin to routing and the
edit-pattern constraint rather than to in-loop gate access.
\subsection{Lift distribution and selection reliability}
\label{sec:results_selection_risk}
\begin{figure*}[t]
    \centering
    \includegraphics[width=0.62\textwidth]{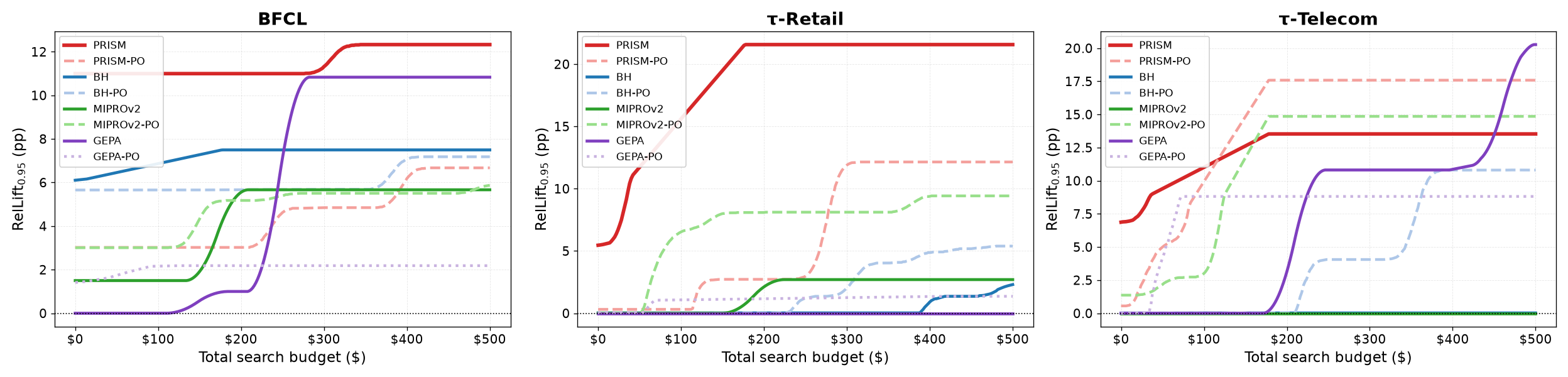}
    \caption{
Reliable lift as a function of search budget. Each curve answers: given
this budget, if a deployer ran as many optimizer runs as it affords and
picked one using only pre-scorecard information, how much held-out lift
would the picked harness deliver in the unlucky (5th-percentile) case?
Solid curves are \textsc{-MW} variants, dashed are \textsc{-PO}. At the
vertical line (\(B_{\mathrm{ref}}\)), curve heights equal the
\(\mathrm{RelLift}_{95}\) column of
Table~\ref{tab:main-scorecard-results}. Budgets use the inner-rollout cost
convention of Appendix~\ref{app:cost-methodology}; the curves compare
selection reliability across optimizers, not deployment economics.
}
    \label{fig:budget_curves}
\end{figure*}
Because the optimizers are stochastic, peak discovery and selected-harness
reliability differ. \textsc{MaxLift} measures whether search ever finds a
strong harness; \textsc{WorstLift}, \(\mathrm{RR}_{0}\), and
\(\mathrm{RelLift}_{95}\) measure whether the optimizer reliably returns and
selects one. Here and in Section~\ref{sec:ablation}, \(\mathrm{RelLift}_{95}\) values
are empirical lower-tail estimates from each arm's observed runs
(\S\ref{sec:budgeted-estimator}), read alongside \(\mathrm{WorstLift}\) and
\(\mathrm{RR}_0\), not guarantees; estimator-level confidence intervals are
reported for the \textsc{PRISM-MW} arms in Appendix~\ref{app:subsampling}, and
cross-optimizer \(\mathrm{RelLift}_{95}\) comparisons rest on point
estimates.
On BFCL, all methods reach \(\mathrm{RR}_{0}=100\%\), so the comparison is
magnitude: PRISM leads \textsc{MeanLift} (14.2 pp), \textsc{WorstLift}
(11.0 pp), and \(\mathrm{RelLift}_{95}\) (12.0 pp). The \(\tau^2\) tasks
expose selection risk. BH finds the best individual harnesses on both, but
with \textsc{WorstLift}=\(-13.5\) pp and negative
\(\mathrm{RelLift}_{95}\): a strong explorer, an unreliable selector.
PRISM's lower peak, by contrast, comes with a positive selected tail
(\(\mathrm{RelLift}_{95}=10.1\) pp on Retail, 7.8 pp on Telecom). The main
result is therefore not higher average lift alone but a more stable
selected-harness distribution under pre-scorecard selection.
\subsection{Budgeted reliable lift}
\label{sec:results_budget}
Figure~\ref{fig:budget_curves} shows the budgeted version of the same
selection problem; since selection uses pre-scorecard statistics, the curve is
not an oracle best-of-\(N\) score. On BFCL, PRISM is highest across the
logged budget range. On \(\tau^2\)-Retail, PRISM rises early and plateaus
highest; GEPA is cheap but saturates low, while BH/BH-PO show upside without
reliably converting budget into deployable selected lift. On
\(\tau^2\)-Telecom, PRISM gives strong low-budget reliability, but PRISM-PO
and MIPROv2 can reach higher plateaus with larger budgets; efficiency should
therefore be read from the budget curve together with the accounting
convention, not from \textsc{Cost/run} or \textsc{MaxLift} alone.
\subsection{Prompt versus prompt-plus-middleware}
\label{sec:results_prompt_vs_middleware}
The comparison between prompt and prompt-plus-middleware variants tests whether
gains come from better instructions alone or from exposing runtime behavior as
an optimization surface. Appendix~\ref{app:examples} gives concrete
benchmark-level examples of the edits discussed here, including BFCL schema
repairs, \(\tau^2\)-Retail state-consistency guards, and a \(\tau^2\)-Telecom
prompt-only policy checklist.
Prompt-only search is useful: both PRISM-PO and BH-PO improve over baseline in
several settings. However, full PRISM has stronger lower-tail reliability,
suggesting that some failures are not instruction-local. The examples in
Appendices~\ref{app:examples-bfcl} and~\ref{app:examples-tau-retail} illustrate
such cases: malformed arguments, wrapper mismatches, missing preconditions, and
state-consistency violations can often be checked or normalized at the tool
boundary via the three edit patterns of \S\ref{sec:optimizers}: silent
correction of malformed arguments, error blocking so the model retries, and
prerequisite blocking before the environment is mutated.
The BH results show an important caveat: middleware is not automatically
beneficial. On \(\tau^2\)-Retail, BH-PO outperforms BH in mean lift and
lower-tail reliability. This benchmark contains one-shot mutations, mutable
order state, and strict policy constraints, so an unconstrained middleware edit
can introduce brittle runtime behavior or change the semantics of a tool call.
In such cases, adding middleware increases the search space but also increases
selection risk; by contrast, the pattern-guided surface lifts every optimizer
that receives it (\S\ref{sec:results}), so the constraint,
not the surface alone, is what makes middleware safely searchable.
PRISM differs by treating middleware as a routed repair surface rather than a
generic place to edit. Middleware is most useful when the failure is locally
observable and checkable at the tool boundary; prompts remain the more
appropriate surface for global task policy, such as what information to seek,
how to follow service policy, or when to escalate. The
\(\tau^2\)-Telecom example in Appendix~\ref{app:examples-tau-telecom} illustrates
this distinction: the main fix is a prompt-level policy checklist rather than
argument rewriting.
The discovered middleware rules are domain-specific by design and do not
transfer across benchmarks (the Retail and BFCL rules share zero tool
names, and each would be a no-op in the other domain), but the discovery
procedure does transfer: the same three edit patterns and routing, with no
hand-tuning, re-derived the appropriate rules per domain and, on
\(\tau^2\)-Telecom, correctly decided that the tool boundary was not the
right surface at all, producing no middleware and placing the fix in the
prompt (PRISM-MW still exceeds PRISM-PO on Telecom because the two variants
follow different search trajectories even when the returned middleware is
empty). Transfer beyond conversational tool agents (e.g., code or research
agents) is neither claimed nor tested (see Limitations).
Thus, the result is not that middleware always helps. The stronger claim is
that middleware helps when the optimizer can identify local runtime invariants
and avoid broad, semantics-changing patches. PRISM's advantage comes from
searching both surfaces with failure-localized routing: prompts handle semantic
policy, while middleware handles checkable execution constraints.
\section{Ablation Study}
\label{sec:ablation}
We report two ablations. Section~\ref{sec:ablation-component} ablates
PRISM's components individually, reverting one attribute of
Table~\ref{tab:optimizer_comparison} at a time.
Section~\ref{sec:ablation_inner_model} ablates the two model roles (the
inner model executes the task, the outer model proposes harness edits);
these are sensitivity analyses rather than a full factorial decomposition
of PRISM, with the full grids, run provenance, and cost-scope notes in
Appendix~\ref{app:model-ablations}.
\subsection{Component ablation}
\label{sec:ablation-component}
Each variant reverts exactly one attribute of
Table~\ref{tab:optimizer_comparison} to the baselines' setting, with
everything else held fixed, under the identical protocol and budget and with
all arms sharing one proposer so comparisons are internal:
\textsc{PRISM-NoRoute} computes clusters but does not route fixes to surface
slots; \textsc{PRISM-NoGate} consults the gate at final acceptance only, the
same discipline as GEPA and MIPROv2; \textsc{PRISM-NoMatrix} clusters
current-generation failures only, without cross-iteration tracking;
\textsc{PRISM-NoConstraint} lifts the three-pattern restriction and lets the
middleware mutator edit execution logic freely; and
\textsc{PRISM-NoCrossover} skips the crossover step while keeping the main
runs' acceptance criterion.
Table~\ref{tab:component-ablation} reports the grids on all three tasks;
PRISM-PO is included as a reference point, and
GEPA-PO\textsuperscript{\dag} reference values appear in
Table~\ref{tab:main-scorecard-results}.
\begin{table*}[t]
\centering
\scriptsize
\setlength{\tabcolsep}{4.0pt}
\renewcommand{\arraystretch}{1.0}
\caption{
Component ablation. Each variant switches off exactly one PRISM component
and keeps everything else fixed (\S\ref{sec:ablation-component}); comparing
its row against PRISM-MW (full) shows what that component contributes.
Lifts are percentage points over \(h_0\).
\textsuperscript{\S}~marks Telecom reference rows that use
Table~\ref{tab:main-scorecard-results}'s post-hoc \(\mathrm{pass}^{4}\)
scoring while the ablation arms use the native protocol, so Telecom
comparisons against those rows carry this caveat (see Limitations).
}
\label{tab:component-ablation}
\begin{tabular}{@{} l l l r l r r r @{}}
\toprule
Task & Variant & Reverts & MeanLift & 95\% CI & WorstLift & \(\mathrm{RR}_{0}\) & \(\mathrm{RelLift}_{95}\) \\
\midrule
\multirow{7}{*}{\(\tau^2\)-Retail}
& PRISM-PO            & middleware surface   & 1.8  & [-9.5, 12.2]  & -9.5  & 67\%  & -5.4 \\
& PRISM-NoRoute       & routing \(\to\) none & 5.4 & [0.1, 7.7]  & -1.2  & 87\%  & 1.2  \\
& PRISM-NoGate        & gate use \(\to\) final only & 12.9 & [7.3, 18.5]   & 6.3   & 100\% & 9.5  \\
& PRISM-NoMatrix      & failure matrix \(\to\) no & 12.8 & [7.3, 19.7]   & 5.1   & 100\% & 9.1  \\
& PRISM-NoConstraint  & edits \(\to\) unconstrained & 4.7  & [-3.3, 12.7]  & -4.7  & 75\%  & 2.7  \\
& PRISM-NoCrossover   & crossover \(\to\) off & 4.6 & [1.6, 7.7] & -1.4 & 81\% & 0.3 \\
& PRISM-MW (full)     & ---                  & \textbf{14.9} & [8.4, 19.9] & \textbf{5.4} & \textbf{100\%} & \textbf{10.1} \\
\midrule
\multirow{7}{*}{BFCL}
& PRISM-PO            & middleware surface   & 5.8  & [3.9, 7.7]    & 3.0   & 100\% & 3.9  \\
& PRISM-NoRoute       & routing \(\to\) none & 5.3 & [0.7, 9.7]  & 0.7   & 100\% & 5.5  \\
& PRISM-NoGate        & gate use \(\to\) final only & 10.7 & [7.1, 14.3]   & 6.3   & 100\% & 9.9  \\
& PRISM-NoMatrix      & failure matrix \(\to\) no & 10.0 & [6.5, 13.5]   & 5.6   & 100\% & 10.5 \\
& PRISM-NoConstraint  & edits \(\to\) unconstrained & 6.5  & [1.1, 11.9]   & 0.1   & 100\% & 5.6  \\
& PRISM-NoCrossover   & crossover \(\to\) off & 11.8 & [9.5, 14.1] & 5.8 & 100\% & 8.8 \\
& PRISM-MW (full)     & ---                  & \textbf{14.2} & [11.7, 16.7] & \textbf{11.0} & \textbf{100\%} & \textbf{12.0} \\
\midrule
\multirow{7}{*}{\(\tau^2\)-Telecom}
& PRISM-PO\textsuperscript{\S}            & middleware surface   & 6.8  & [1.4, 14.2]   & 0.0   & 75\%  & 2.0  \\
& PRISM-NoRoute       & routing \(\to\) none & 2.4  & [0.3, 5.4]   & 0.0   & 75\%  & 1.4  \\
& PRISM-NoGate        & gate use \(\to\) final only & 1.4  & [-2.4, 4.4]  & -4.1  & 75\%  & 0.9   \\
& PRISM-NoMatrix      & failure matrix \(\to\) no & -0.3 & [-3.0, 2.7]  & -4.1  & 25\%  & 0.1   \\
& PRISM-NoConstraint  & edits \(\to\) unconstrained & 0.7  & [-1.7, 3.0]  & -2.7  & 50\%  & 0.5   \\
& PRISM-NoCrossover   & crossover \(\to\) off & 2.7 & [-1.9, 9.2] & -2.3 & 75\% & 1.6 \\
& PRISM-MW (full)\textsuperscript{\S}     & ---                  & \textbf{10.1} & [7.4, 12.8]   & \textbf{6.8} & \textbf{100\%} & \textbf{7.8} \\
\bottomrule
\end{tabular}
\end{table*}
Table~\ref{tab:component-ablation} yields three takeaways. First,
\emph{routing and the pattern constraint are the only components whose
removal is first-order on all three tasks}: removing either collapses both
mean lift and the reliability tail, with the \textsc{NoRoute} interval
disjoint from full PRISM-MW's on Retail and BFCL. Second, gate access and
the failure matrix are second-order on Retail and BFCL; the NoGate result
also answers the fairness question of \S\ref{sec:run-protocol} by
measurement, since PRISM under the baselines' own gate discipline still
exceeds the strongest same-surface baseline on Retail.
Third, \(\tau^2\)-Telecom departs from this pattern: every removal is
first-order there, though the Telecom reference rows are post-hoc
\(\mathrm{pass}^{4}\)-scored while the ablation arms use the native path,
so the uniformity may partly reflect that orchestration difference.
What does \emph{not} produce the gain: clustering alone
(\textsc{PRISM-NoRoute} and \textsc{PRISM-PO} both retain it), and surface
access alone (GEPA-MW and MIPROv2-MW receive PRISM's exact middleware
surface and do not close the gap). Among inherited components, crossover is
second-order on BFCL but first-order on both \(\tau^2\) domains, where its
removal erases most of the mean lift and, on Retail, nearly all of
\(\mathrm{RelLift}_{95}\). Frontier retention is bounded indirectly:
GEPA-MW already combines Pareto retention with PRISM's middleware surface
yet trails full PRISM.
\subsection{Model-role sensitivity}
\label{sec:ablation_inner_model}
We ablate the two model roles; the full grids are in
Appendix~\ref{app:model-ablations}
(Tables~\ref{tab:inner-model-ablation-bfcl-retail}
and~\ref{tab:outer-model-ablation-bfcl-retail}).
\paragraph{Inner model.}
Table~\ref{tab:inner-model-ablation-bfcl-retail} shows that the strongest
absolute execution model is not always the best optimization target. On BFCL,
GPT-5.4 reaches the highest absolute score (\(\mathrm{MaxSC}\) 73.6) and the
largest single-run lift, but GPT-5.4-mini matches its mean lift with a
stronger lower tail at roughly a fifth of the logged cost. On
\(\tau^2\)-Retail, GPT-5.4-mini leads every lift and reliability statistic,
while GPT-5.4 attains the highest absolute gate score (\(\mathrm{MaxSC}\)
59.0). Weaker inner models
occasionally find positive candidates, but their lower-tail metrics are often
near zero or negative, suggesting that harness search needs a minimum level of
tool-use competence to produce reliably selectable updates.
\paragraph{Outer model.}
Table~\ref{tab:outer-model-ablation-bfcl-retail} shows that the proposer
changes both repair quality and selection risk. On BFCL, Opus 4.7 is strongest
across mean, maximum, worst-case, and reliable lift, consistent with BFCL's
dependence on precise schema and middleware repair. On \(\tau^2\)-Retail,
quality is less monotonic: Sonnet 4.6 gives the largest mean and max lift,
while GPT-5.5 gives the strongest lower-tail reliability; cheaper proposers can
also work when their edits match the benchmark's repair structure. The practical
interpretation is risk-dependent: lower-cost proposers are useful for
exploration, but stronger proposers are safer when the selected harness must be
reliably better rather than occasionally better.
\section{Conclusion}
\label{sec:conclusion}
We studied automatic harness optimization for fixed-model tool agents. Our
contributions are a reliability-centered evaluation protocol for budgeted
harness selection; \(\mathrm{RelLift}_{95}\), its budgeted reliable-lift
metric; and PRISM, a failure-routed evolutionary optimizer over factorized
prompt and tool-boundary middleware edits. Across BFCL multi-round,
\(\tau^2\)-Retail, and \(\tau^2\)-Telecom, PRISM shows the strongest
selected-harness profile in our main comparison: the highest mean held-out
lift and positive \(\mathrm{RelLift}_{95}\) on all three benchmarks, with
the component ablation attributing this margin chiefly to failure-surface
routing and the edit-pattern constraint, the only two components whose
removal is first-order on all three tasks, rather than to in-loop gate
access or the failure matrix. The empirical claim is not that PRISM
guarantees improvement on every future repair distribution, but that under
a fixed rollout budget and scorecard-held-out protocol, failure-routed
prompt/middleware search produces a more reliably selectable harness than
the compared prompt-only and hill-climbing baselines.
\section*{Limitations}
\textbf{Benchmark scope.} Our evaluation covers three multi-turn tool-use
benchmarks: BFCL multi-round, \(\tau^2\)-Retail, and \(\tau^2\)-Telecom.
They span function calling, simulated service workflows, policy following,
and tool-mediated state changes, but remain controlled offline settings:
they do not capture production traffic, changing user populations,
long-running memory, adversarial users, or business-specific safety
policies. We also do not demonstrate a full multi-cycle deployment loop; no
experiment re-optimizes against failures produced by a previously optimized
harness, and using reliability as an in-loop stopping signal is future
work.

\textbf{Seed harnesses.} Our baselines are strong human harnesses: the
\(\tau^2\) prompts are the benchmarks' official hand-written policy
documents, and the middleware repairs target failure modes largely
orthogonal to prompt quality. The fully controlled test, seeding the search
with the best prompt-only-optimized harness, remains open.

\textbf{Scorecard size and simulator stochasticity.} The \(\tau^2\)-bench
scorecards are small, held-out subsets of the public task sets
(\S\ref{sec:run-protocol}) and use stochastic user simulation. Scoring each
case over four rollouts limits the influence of any single trial, but
\(\mathrm{pass}^{4}\) is a stricter metric, not merely a less noisy one,
and the table intervals and \(\mathrm{RelLift}_{95}\) values are empirical
estimates from the observed runs, not guarantees. The resampling procedures
summarize optimizer-run variability under our fixed protocol only;
Appendix~\ref{app:subsampling} reports how the \textsc{PRISM-MW} estimates
move as runs are subsampled. Broader validation on more domains, larger
scorecards, more split seeds, and live deployments would strengthen
external validity.

\textbf{Telecom scoring caveat.} On \(\tau^2\)-Telecom, the
\textsc{PRISM-PO} and \textsc{PRISM-MW} results (marked
\textsuperscript{\S}) were scored from four genuine simulator trials per
case (one original plus three run afterwards), AND-combined outside the
benchmark-native \(\mathrm{pass}^{4}\) path used by the other Telecom arms,
so cross-arm Telecom comparisons carry this orchestration caveat.

\textbf{Cost accounting.} Our costs compare optimizer search overhead under
our implementation; the controlled unit is the inner-model rollout budget,
and the monetary columns follow the convention of
Appendix~\ref{app:cost-methodology} with final scorecard evaluation
excluded. They are implementation-level diagnostics, not an end-to-end
economic analysis or provider-independent pricing.

\textbf{Middleware risk.} Middleware as an optimization surface increases
both power and risk: poorly constrained edits can change tool semantics,
and lifting the pattern constraint sharply degrades reliability
(\S\ref{sec:ablation-component}). Our results support middleware
optimization only where failures are locally checkable at the tool
boundary. The three-pattern vocabulary excludes repairs requiring
cross-call coordination, retries, asynchronous state, or control-flow
changes; richer middleware programs under the same routed-and-gated
discipline, and extension to code or research agents, are future work. In
deployment, PRISM-generated middleware should be reviewed, tested, and
versioned before release.
\section*{Ethics Statement}
The experiments use public benchmark artifacts and simulated service-agent
environments rather than newly collected human-subject data. We did not recruit
annotators, collect user data, or add personally identifying information. The
service records used in the \(\tau^2\)-bench environments, such as order IDs,
item IDs, account states, and tool outputs, are benchmark-provided simulation
artifacts. We report aggregate scores and selected qualitative examples only to
illustrate failure modes. We do not release raw trajectories containing complete
simulated account histories unless permitted by the benchmark license.
\section*{Acknowledgements}
We thank the anonymous reviewers and the area chair, whose feedback
substantially shaped the final version, in particular the component ablation
and the expanded evaluation protocol. We thank the Airbnb ML infrastructure
team for supporting the Azure OpenAI and Bedrock evaluation capacity used in
these experiments, and our leadership, Ben Loulier, Yi Li, David Haggerty,
and Ahmad Al-Dahle, for their support of this work. An AI assistant was used for writing and phrasing assistance and for coding assistance during development of the released codebase; all text and code were reviewed and approved by the authors.
\bibliography{main}
\appendix
\section{Model-Role Ablation Tables}
\label{app:model-ablations}
Tables~\ref{tab:inner-model-ablation-bfcl-retail}
and~\ref{tab:outer-model-ablation-bfcl-retail} report the full inner- and
outer-model sensitivity grids discussed in
Section~\ref{sec:ablation_inner_model}.
Both grids are independent reruns rather than the same run tuples as
Table~\ref{tab:main-scorecard-results}, so we treat them as additional
evidence rather than replacements for the main comparison. Costs follow
Appendix~\ref{app:cost-methodology} and are logged search diagnostics;
these tables report Cost/run (all), over all search-time model calls
including outer-proposer calls, so the main configuration's figure here is
higher than Table~\ref{tab:main-scorecard-results}'s Cost/run (inner).
\begin{table*}[t]
\centering
\scriptsize
\setlength{\tabcolsep}{3.6pt}
\renewcommand{\arraystretch}{1.08}
\caption{
How PRISM's results change with the inner model (the model that executes
the task). The outer proposal model is fixed to Opus 4.7. Lift columns are
defined as in Table~\ref{tab:main-scorecard-results};
\(\mathrm{MaxSC}\) is the best absolute gate-split pass rate
(single-trial) reached during search. Cost/run (all) prices all
search-time model calls, including outer-proposer calls.
}
\label{tab:inner-model-ablation-bfcl-retail}
\resizebox{\textwidth}{!}{
\begin{tabular}{
@{}
p{0.12\textwidth}
l
r
l
r
r
r
r
r
r
@{}
}
\toprule
Benchmark
& Inner model
& MeanLift
& 95\% CI
& MaxLift
& \(\mathrm{MaxSC}\)
& WorstLift
& \(\mathrm{RelLift}_{95}\)
& \(\mathrm{RR}_{0}\)
& Cost/run (all) \\
\midrule
\multirow{5}{*}{\shortstack[l]{BFCL\\600-case\\\(\mathrm{pass}^{1}\)}}
& GPT-5.4-mini  & \textbf{12.1} & [10.0, 14.6]  & 15.8 & 66.9 & \textbf{9.2} & \textbf{10.1} & \textbf{100\%} & \$126 \\
& GPT-5.4       & 12.0          & [7.2, 16.8]   & \textbf{16.8} & \textbf{73.6} & 5.2 & 8.0 & \textbf{100\%} & \$582 \\
& Qwen3-32B     & -1.8          & [-7.0, 4.0]   & 4.0  & 52.7 & -7.0  & -5.4  & 30\% & \textbf{\$44} \\
& GPT-5nano     & -12.8         & [-35.7, -0.2] & -0.2 & 60.0 & -35.7 & -24.7 & 0\%  & \$48 \\
& GPT-OSS-20B   & 5.2           & [-0.3, 12.5]  & 12.5 & 58.0 & -0.3  & 0.9   & 70\% & \$45 \\
\midrule
\multirow{5}{*}{\shortstack[l]{\(\tau^2\)-Retail\\74-case\\\(\mathrm{pass}^{4}\)}}
& GPT-5.4-mini  & \textbf{21.3} & [15.2, 27.0] & \textbf{29.7} & 50.0 & \textbf{12.2} & \textbf{15.5} & \textbf{100\%} & \$16 \\
& GPT-5.4       & 3.4           & [-4.7, 10.1] & 10.8 & \textbf{59.0} & -9.5  & -4.4 & 80\% & \$16 \\
& Qwen3-32B     & 0.3           & [-6.1, 5.4]  & 6.8  & 30.3 & -9.5  & -4.7 & 50\% & \$13 \\
& GPT-5nano     & 0.7           & [-6.8, 8.1]  & 8.1  & 38.2 & -6.8  & -6.8 & 50\% & \$17 \\
& GPT-OSS-20B   & -0.9          & [-4.0, 2.7]  & 2.7  & 26.1 & -4.0  & -3.1 & 30\% & \textbf{\$11} \\
\bottomrule
\end{tabular}
}
\end{table*}
\begin{table*}[t]
\centering
\scriptsize
\setlength{\tabcolsep}{4.0pt}
\renewcommand{\arraystretch}{1.08}
\caption{
How PRISM's results change with the outer model (the model that proposes
harness edits). The inner execution model is fixed to GPT-5.4-mini.
Columns are defined as in
Table~\ref{tab:inner-model-ablation-bfcl-retail}.
}
\label{tab:outer-model-ablation-bfcl-retail}
\resizebox{\textwidth}{!}{
\begin{tabular}{
@{}
p{0.12\textwidth}
l
r
l
r
r
r
r
r
r
@{}
}
\toprule
Benchmark
& Outer model
& MeanLift
& 95\% CI
& MaxLift
& \(\mathrm{MaxSC}\)
& WorstLift
& \(\mathrm{RelLift}_{95}\)
& \(\mathrm{RR}_{0}\)
& Cost/run (all) \\
\midrule
\multirow{6}{*}{\shortstack[l]{BFCL\\600-case\\\(\mathrm{pass}^{1}\)}}
& Opus 4.7      & \textbf{14.2} & [11.7, 16.7] & \textbf{17.5} & \textbf{68.6} & \textbf{11.0} & \textbf{12.0} & \textbf{100\%} & \$142 \\
& Sonnet 4.6    & 9.7           & [8.0, 11.2]  & 11.8 & 62.9 & 7.2  & 8.2  & \textbf{100\%} & \$69 \\
& GPT-5.5       & 9.8           & [7.1, 12.6]  & 12.8 & 63.9 & 6.5  & 7.4  & \textbf{100\%} & \$117 \\
& Kimi K2.5     & 2.2           & [-3.2, 7.6]  & 8.5  & 59.6 & -4.5 & -2.5 & 50\%  & \$35 \\
& Qwen3-Coder   & 2.0           & [-0.8, 4.8]  & 6.0  & 57.1 & -2.2 & -0.6 & 80\%  & \$24 \\
& GPT-OSS-120B  & -1.5          & [-6.7, 5.0]  & 8.0  & 59.1 & -8.3 & -6.2 & 20\%  & \textbf{\$12} \\
\midrule
\multirow{6}{*}{\shortstack[l]{\(\tau^2\)-Retail\\74-case\\\(\mathrm{pass}^{4}\)}}
& Opus 4.7      & 14.9          & [8.4, 19.9]  & 21.6 & 60.1 & 5.4   & 10.1 & \textbf{100\%} & \$64 \\
& Sonnet 4.6    & \textbf{21.3} & [15.2, 27.0] & \textbf{29.7} & \textbf{68.2} & 12.2 & 15.5 & \textbf{100\%} & \$16 \\
& GPT-5.5       & 19.3          & [16.9, 22.6] & 24.3 & 62.8 & \textbf{16.2} & \textbf{16.9} & \textbf{100\%} & \$45 \\
& Kimi K2.5     & 15.2          & [13.2, 16.9] & 17.6 & 56.1 & 12.2 & 13.5 & \textbf{100\%} & \$2 \\
& Qwen3-Coder   & 3.0           & [-8.8, 12.5] & 14.9 & 53.4 & -13.5 & -6.4 & 80\%  & \textbf{\$1} \\
& GPT-OSS-120B  & 16.7          & [10.8, 27.0] & 27.0 & 65.5 & 10.8 & 11.3 & \textbf{100\%} & \$3 \\
\bottomrule
\end{tabular}
}
\end{table*}
\section{Evaluation Protocol Details}
\label{app:experiment-protocol}
Figure~\ref{fig:experiment-process} gives the full resource-bounded evaluation
procedure used for each benchmark--optimizer pair. The main text summarizes the
protocol; this appendix spells out the repeated-run structure and the bootstrap
estimator for \(\mathrm{RelLift}_{95}(B)\).
\begin{figure*}[t]
\small
\centering
\fbox{
\begin{minipage}{0.96\textwidth}
\textbf{Evaluation protocol for one benchmark--optimizer pair}
\begin{enumerate}
    \item Construct four split seeds. Each seed defines disjoint
    \(D_{\mathrm{repair}}\), \(D_{\mathrm{gate}}\), and
    \(D_{\mathrm{score}}\). The scorecard split is hidden during search.
    \item For each split seed \(s \in \{1,\ldots,4\}\), run up to four
    independent optimizer repeats under the same inner-model rollout budget.
    \item In each optimizer repeat:
    \begin{enumerate}
        \item Search over harness candidates using only
        \(D_{\mathrm{repair}}\) and \(D_{\mathrm{gate}}\).
        \item Enforce a fixed number of inner-model task rollouts during search.
        \item Enforce per-task resource limits on every rollout: timeout,
        maximum turns, and token budget. Resource violations, crashes,
        repeated tool-call loops, and abnormal terminations receive zero score.
        \item Select one final harness using a pre-scorecard statistic
        \(G_i\), computed only from repair/gate-visible information.
    \end{enumerate}
    \item Evaluate the selected harness once on \(D_{\mathrm{score}}\), paired
    with the baseline harness, and record one complete run tuple
    \[
    (C_i, G_i, \Delta_i),
    \]
    where \(C_i\) is pre-scorecard search cost, \(G_i\) is the pre-scorecard
    selection statistic, and \(\Delta_i\) is held-out scorecard lift.
    \item Discard runs failing the reliability screen of
    Appendix~\ref{app:run-config}, leaving the run pool
    \[
    \mathcal I = \{(C_i, G_i, \Delta_i)\}_{i=1}^{n},
    \qquad n \le 16,
    \]
    where 16 is the design maximum of four split seeds times four repeats.
    Use these runs to compute mean lift, confidence intervals, worst lift, and
    repeatability.
    \item To estimate budgeted reliable lift at budget \(B\):
    \begin{enumerate}
        \item Compute mean pre-scorecard cost
        \[
        \bar C = \frac{1}{n}\sum_{i=1}^{n} C_i,
        \qquad
        N_B = \left\lfloor \frac{B}{\bar C} \right\rfloor .
        \]
        Here \(N_B\) is the average number of affordable independent runs.
        \item Repeat for \(j=1,\ldots,5000\):
        \begin{enumerate}
            \item Sample \(N_B\) runs with replacement from \(\mathcal I\),
            yielding \(\mathcal R_j(B)\).
            \item Select the sampled run with the highest pre-scorecard
            statistic:
            \[
            i_j^\star = \arg\max_{i \in \mathcal R_j(B)} G_i .
            \]
            \item Record the held-out lift of the selected run:
            \[
            z_j = \Delta_{i_j^\star}.
            \]
        \end{enumerate}
        \item Return the lower-tail selected lift:
        \[
        \mathrm{RelLift}_{95}(B)
        =
        \operatorname{Quantile}_{0.05}
        \left(\{z_j\}_{j=1}^{5000}\right).
        \]
    \end{enumerate}
\end{enumerate}
\end{minipage}
}
\caption{
Resource-bounded evaluation protocol. Each benchmark--optimizer pair targets
four split seeds times four independent repeats per seed, a design maximum of
16 runs; the pool actually used is the \(n \le 16\) runs that pass the
reliability screen (Appendix~\ref{app:run-config}). Scorecard outcomes
\(\Delta_i\) are never used during search or
pre-scorecard selection. \(\mathrm{RelLift}_{95}(B)\) is estimated by
bootstrapping budgeted selections from the \(n\) observed runs.
}
\label{fig:experiment-process}
\end{figure*}
\section{PRISM Prompts and Configuration}
\label{app:prism-process}
The generation loop, failure matrix, mutation slots, edit patterns, crossover
rule, and frontier update are described in \S\ref{sec:optimizers}; this
appendix retains prompts and low-level configuration. In the code, the
optimizer's development splits are named train and holdout; in the paper we
refer to them as \(D_{\mathrm{repair}}\) and \(D_{\mathrm{gate}}\), respectively.
The final scorecard split \(D_{\mathrm{score}}\) is never used by PRISM during
analysis, mutation, crossover, frontier update, or pre-scorecard selection.
\paragraph{Analyst prompt.}
The analyst is instructed to group failing cases by root cause rather than by
the last tool call or by superficial API-field differences. The analyst receives
for each failing case a case identifier, stuck type, tool-call sequence, and a
truncated state difference or error message. It outputs JSON clusters with four
fields: root cause, repair surface, case identifiers, and a one-sentence reason
for the surface choice. The repair surface is one of
\texttt{prompt\_only}, \texttt{middleware\_only}, or
\texttt{prompt\_middleware\_both}. The prompt asks for exactly four clusters,
ordered by descending case count, prefers single-surface assignments when
possible, and skips infrastructure-error cases.
\paragraph{Mutation slots.}
Table~\ref{tab:prism-slots} summarizes the three surface-constrained mutation
slots that receive the routed clusters.
\begin{table*}[t]
\centering
\small
\setlength{\tabcolsep}{4pt}
\renewcommand{\arraystretch}{1.08}
\caption{PRISM's three mutation slots. Each generation runs all three in
parallel; the routing of \S\ref{sec:optimizers} decides which failure
clusters each slot receives. PRISM-PO forces all three slots to prompt-only
edits and therefore never modifies middleware.}
\label{tab:prism-slots}
\begin{tabular}{@{}p{0.08\textwidth}p{0.19\textwidth}p{0.25\textwidth}p{0.40\textwidth}@{}}
\toprule
Slot & Variant & Target pattern & Access rule \\
\midrule
p0 & Prompt + middleware & All failures, including joint clusters
& May edit both \texttt{system\_prompt.txt} and middleware. It applies the
single highest-value safe fix across either surface. \\
p1 & Prompt only & Case identifiers from prompt-routed clusters; if no such
cluster is available, PRISM still runs this slot for surface diversity
& May edit only \texttt{system\_prompt.txt}. It writes universal behavioral
rules and adds at most one or two sentences. \\
p2 & Middleware only & Case identifiers from middleware-routed clusters, when such
clusters exist and mutation slots remain
& May edit only middleware. It can silently correct tool-call arguments, block a
call with an error so the model retries, or enforce prerequisite checks before a
tool call. \\
\bottomrule
\end{tabular}
\end{table*}
\paragraph{Mutator workspace and safety checks.}
Before mutation, PRISM writes a workspace containing the current prompt and
middleware, structured failure analysis, a summary of frontier candidates,
history of accepted and rejected candidates, a cross-generation failure matrix,
target failure trajectories, passing trajectories for regression checks, a task
file, a proposal template, and a surface manifest describing which files the
mutator may edit. Mutators are instructed to prioritize recurring cases, skip
persistent cases unless they have a genuinely new angle, inspect passing cases
for safety, never regress fixed cases, and never repeat a rejected approach
without a new angle.
\paragraph{Frontier update details.}
Candidate \(a\) dominates candidate \(b\) only if
\[
S_{\mathrm{gate}}(a) \geq S_{\mathrm{gate}}(b), \quad
R(a) \geq R(b),
\]
and at least one inequality is strict, where \(R=1-\mathrm{Stuck}\). The repair
split score is deliberately excluded from dominance so that repair--gate
tradeoffs do not discard candidates that generalize better. If the frontier
exceeds the population cap, PRISM keeps the highest candidates by
\(S_{\mathrm{gate}}+R\). The final pre-scorecard selection statistic used in
Section~\ref{sec:budgeted-estimator} is the same pair,
\((S_{\mathrm{gate}}, R)\), with deterministic tie-breaking by run identifier.
\section{Prompt and Middleware Examples}
\label{app:examples}
This appendix gives one
representative example per benchmark to illustrate when the useful edit is
middleware-only, prompt-plus-middleware, or prompt-only.
\subsection{BFCL Example}
\label{app:examples-bfcl}
BFCL evaluates exact tool-call correctness, including function names, argument
names, argument types, and JSON structure. This makes it a useful
case for separating prompt repair from middleware repair.
\begin{figure}[t]
    \centering
    \includegraphics[width=\linewidth]{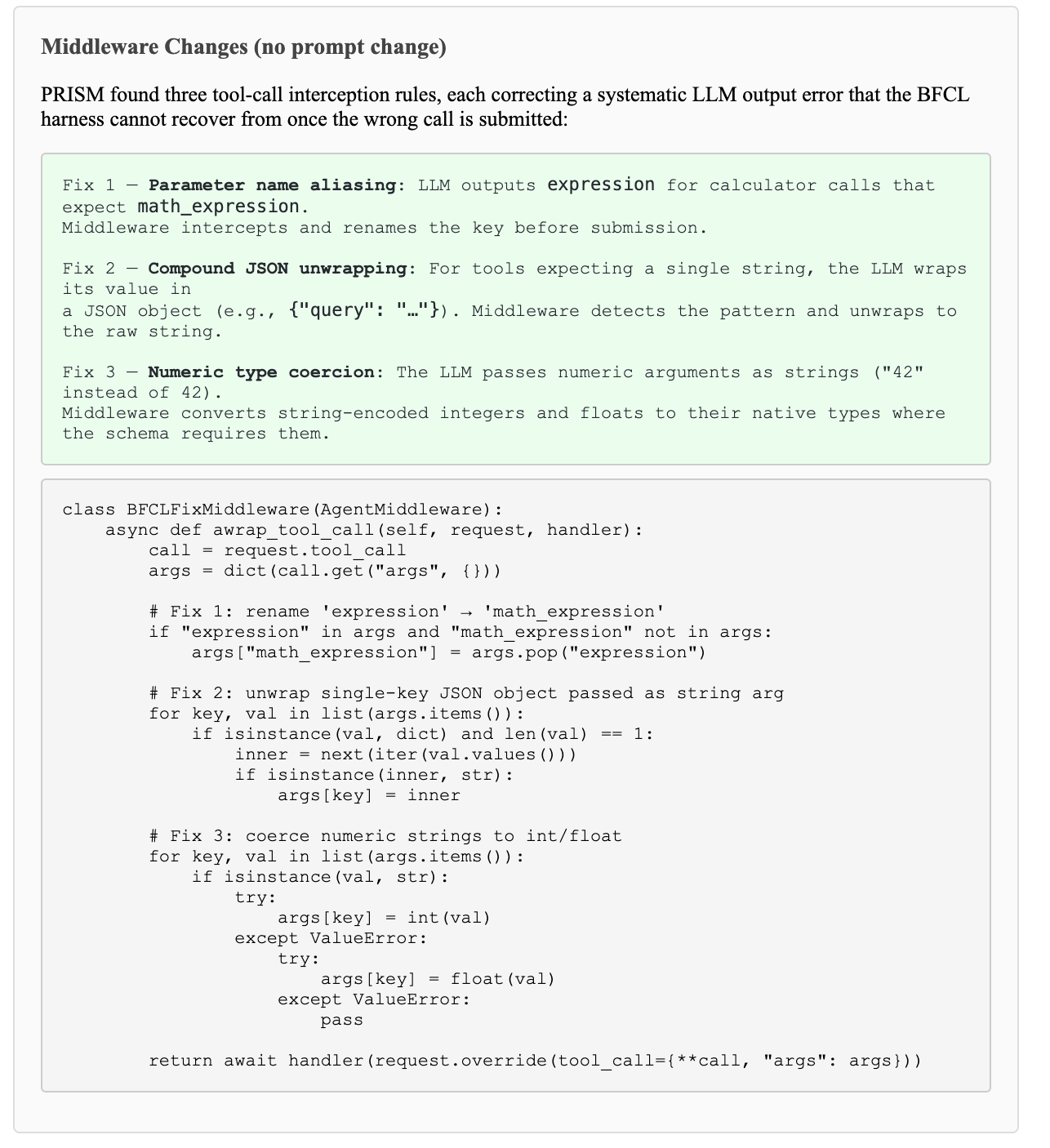}
    \caption{
    BFCL tool-call errors corrected by PRISM middleware.
    }
    \label{fig:bfcl-prism-rules}
\end{figure}
Figure~\ref{fig:bfcl-prism-rules} shows three interception rules found by
PRISM. Each rule repairs a recurring LLM output error before the call
reaches the target function. For example, the model may emit a plausible but
wrong argument name, such as \texttt{expression} instead of
\texttt{math\_expression}. Prompting can reduce this error, but middleware
makes the repair deterministic by rewriting the malformed call into the
expected schema.
\begin{figure}[t]
    \centering
    \includegraphics[width=\linewidth]{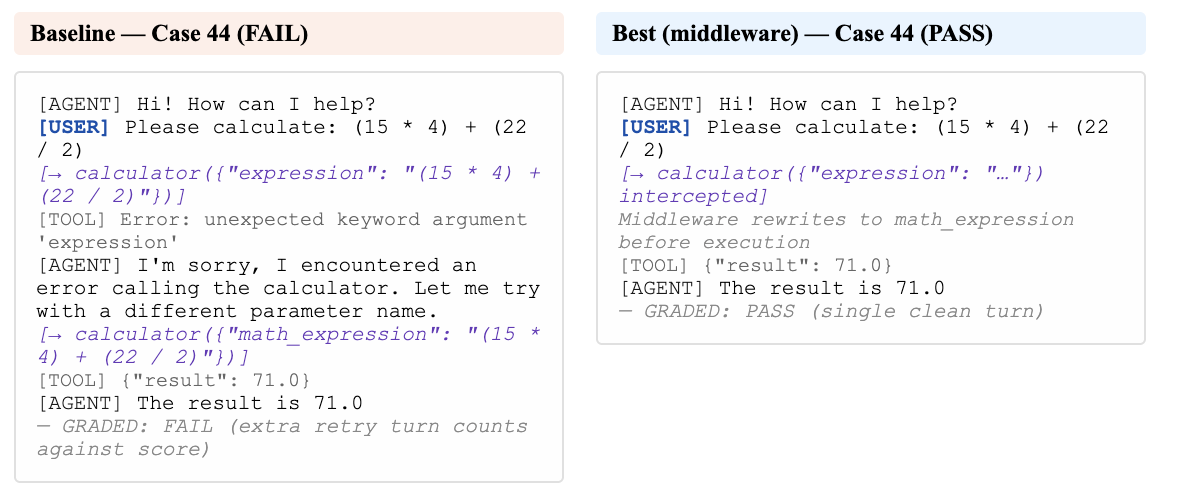}
    \caption{
    Prompt-only repair versus middleware repair on BFCL.
    }
    \label{fig:bfcl-prism-prompt-vs-middleware}
\end{figure}
Figure~\ref{fig:bfcl-prism-prompt-vs-middleware} illustrates the same pattern
for type coercion and JSON unwrapping: these are exact structural repairs, not
only instruction-following problems. This explains why prompt-only
PRISM-PO improves BFCL by \(+5.8\) percentage points, while full
PRISM with middleware improves by \(+14.2\) percentage points
(Table~\ref{tab:main-scorecard-results}).
\subsection{\texorpdfstring{\(\tau^2\)-Retail}{tau2-Retail} Example}
\label{app:examples-tau-retail}
\begin{figure}[t]
    \centering
    \includegraphics[width=\linewidth]{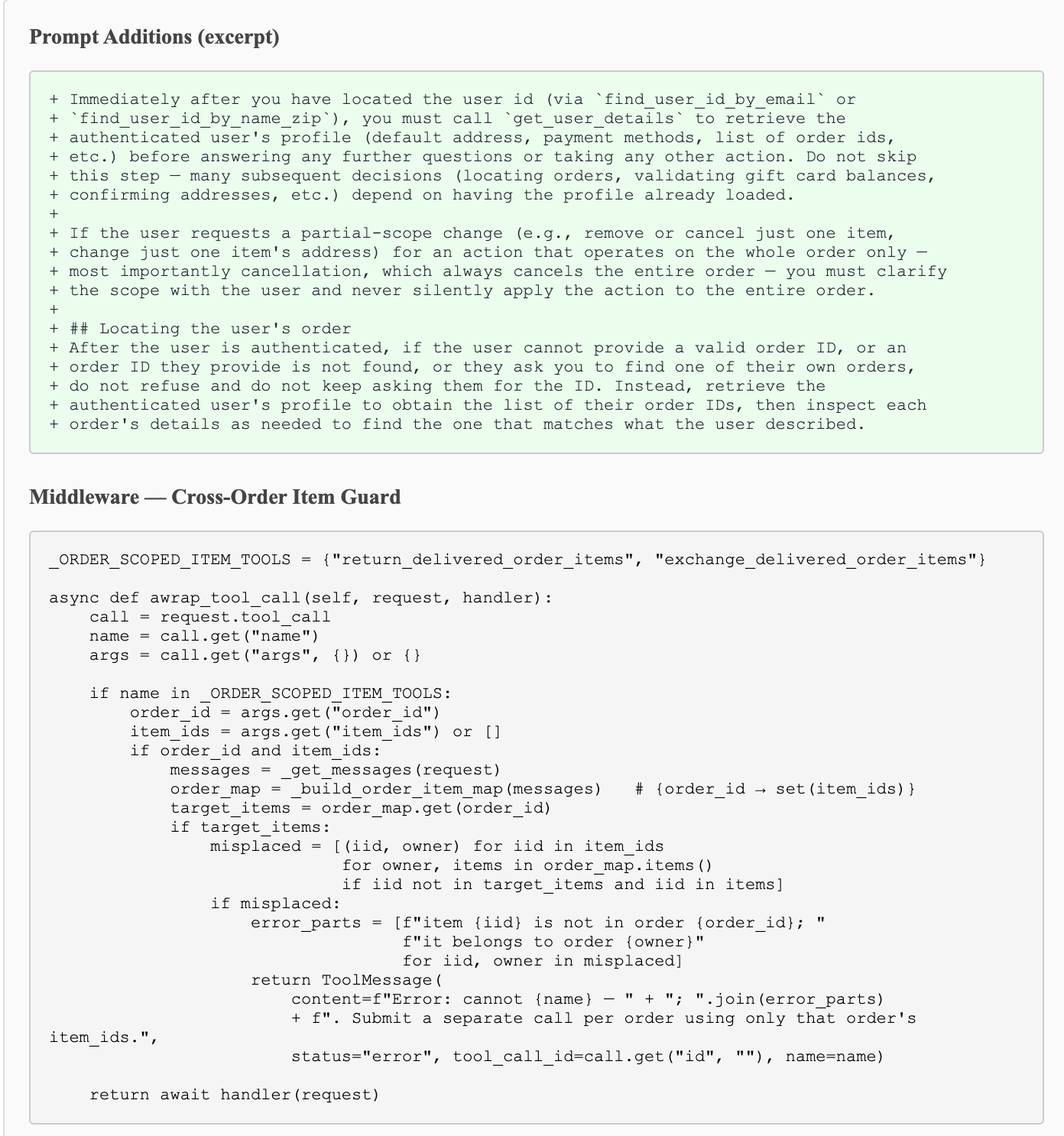}
    \caption{
    Proactive lookup rule found by PRISM on \(\tau^2\)-Retail.
    }
    \label{fig:tau-retail-proactive-lookup}
\end{figure}

\begin{figure}[t]
    \centering
    \includegraphics[width=\linewidth]{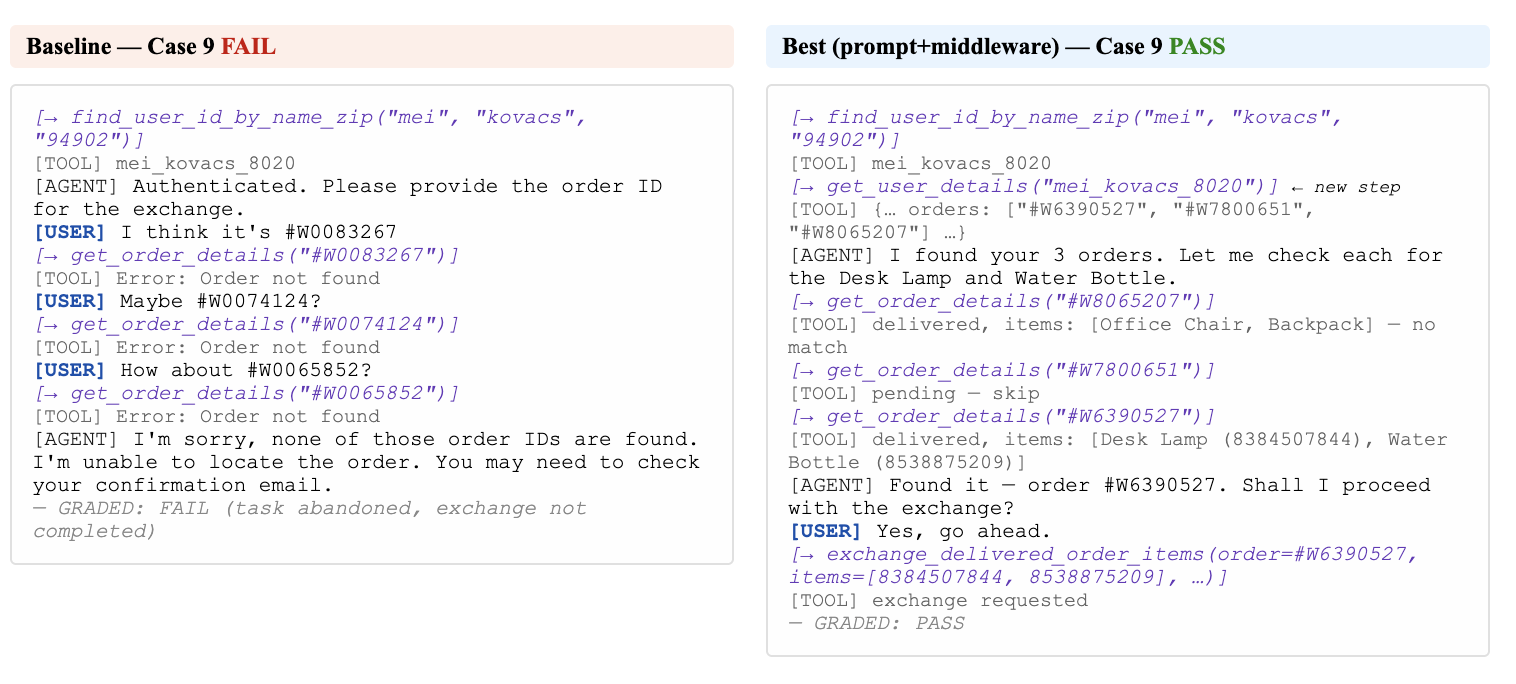}
    \caption{
    Middleware guard for cross-order item validation on \(\tau^2\)-Retail.
    }
    \label{fig:tau-retail-cross-order-guard}
\end{figure}

Figure~\ref{fig:tau-retail-proactive-lookup} shows a proactive-lookup rule for
cases where the user does not know the order ID. In seed 1, every
prompt-only PRISM-PO run remains at 7/20 holdout cases, matching the
baseline. Full PRISM finds this rule within the same search setting by
searching the combined prompt-and-middleware space.
Figure~\ref{fig:tau-retail-cross-order-guard} shows a separate middleware fix
for cross-order item confusion. When a user has multiple delivered orders, the
model can attach an item ID to the wrong order, for example passing item
\texttt{5996159312} from order \texttt{\#W4689314} in a call targeting
\texttt{\#W8855135}. The backend returns only an opaque error, which often
leads to repeated retries. This pattern appears in 8\% of \(\tau^2\)-Retail
baseline trajectories across the tested models.
A prompt can ask the model to verify item--order consistency, but the model only
sees a text summary of prior tool responses. The middleware instead builds a
ground-truth \(\{\text{order} \rightarrow \text{item\_ids}\}\) map from prior
\texttt{get\_order\_details} responses and intercepts invalid calls before
execution. It then returns a precise error identifying the offending item and
its correct order, giving the agent actionable feedback in one step.
\subsection{\texorpdfstring{\(\tau^2\)-Telecom}{tau2-Telecom} Example}
\label{app:examples-tau-telecom}
\begin{figure}[t]
    \centering
    \includegraphics[width=\linewidth]{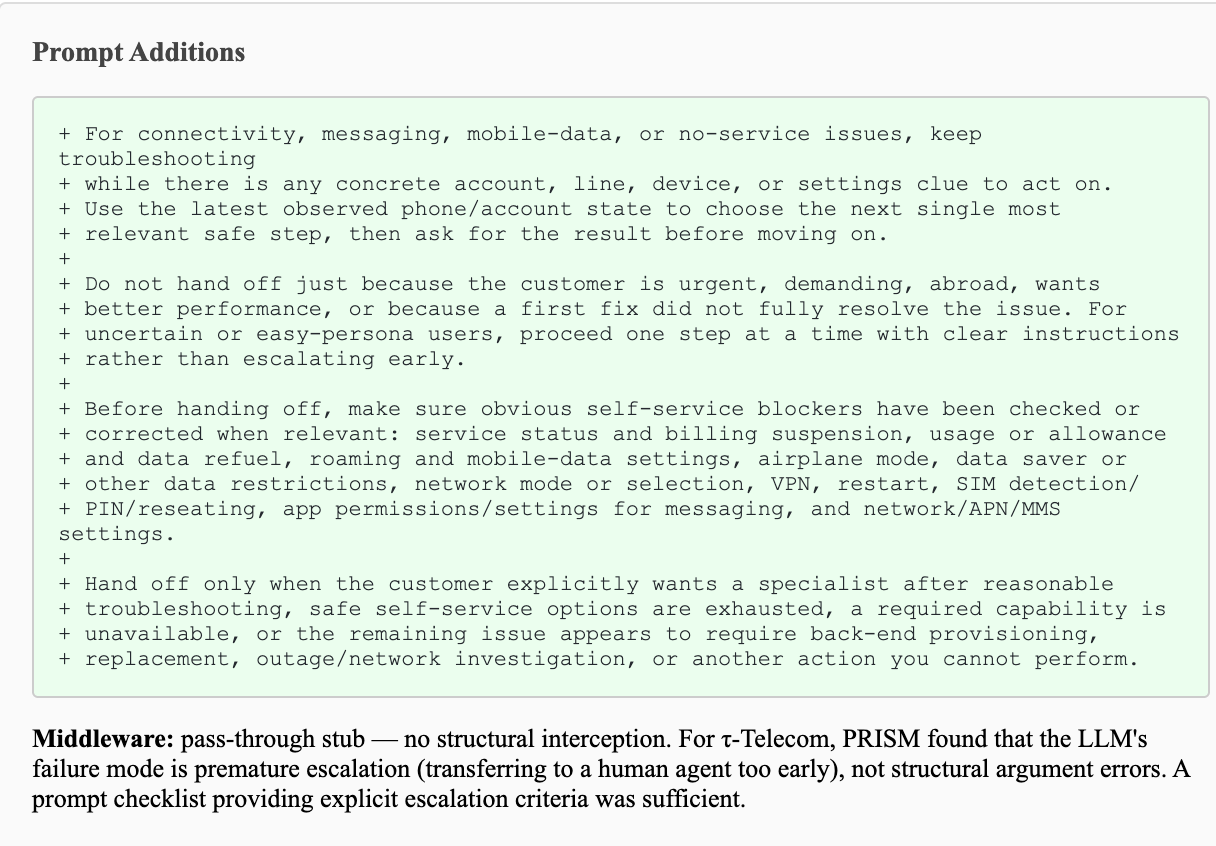}
    \caption{
    Prompt checklist for escalation decisions on \(\tau^2\)-Telecom.
    }
    \label{fig:tau-telecom-checklist}
\end{figure}

\begin{figure}[t]
    \centering
    \includegraphics[width=\linewidth]{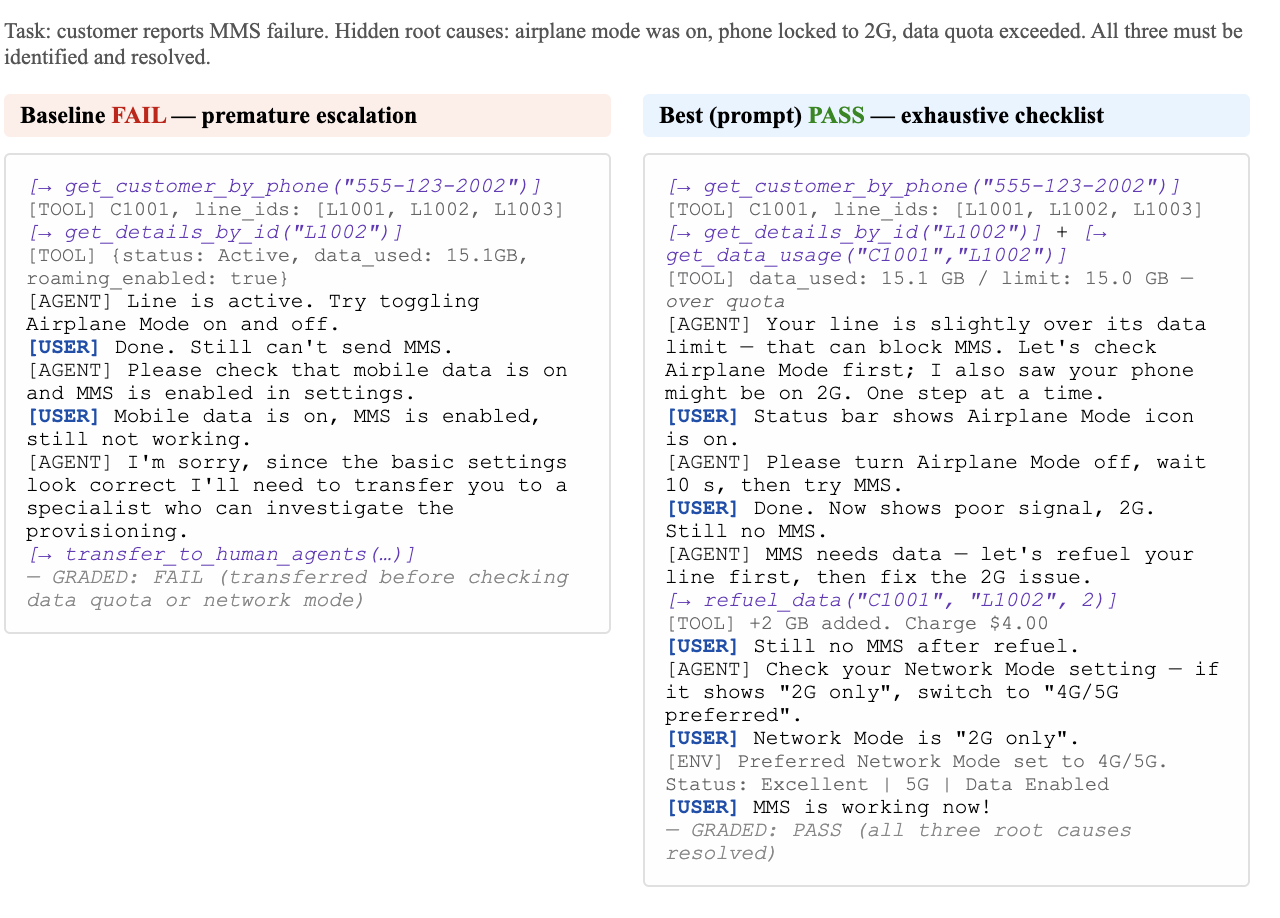}
    \caption{
    Prompt-only repair versus middleware repair on \(\tau^2\)-Telecom.
    }
    \label{fig:tau-telecom-prompt-vs-middleware}
\end{figure}

Figure~\ref{fig:tau-telecom-checklist} shows the main \(\tau^2\)-Telecom fix.
Unlike BFCL or \(\tau^2\)-Retail, the failure is not a structural tool-call error:
the model already knows how to call \texttt{refuel\_data},
\texttt{get\_data\_usage}, and the device/line lookup tools. The missing piece
is an ordered policy checklist for when to continue troubleshooting versus when
to escalate. Adding this checklist to the prompt directly addresses the failure
mode, and the middleware stub confirms that no argument rewriting was needed.
Figure~\ref{fig:tau-telecom-prompt-vs-middleware} contrasts this with the other
tasks. BFCL requires schema-level repair, such as correcting argument names or
types. \(\tau^2\)-Retail requires order-discovery logic and item--order validation
under ambiguity. \(\tau^2\)-Telecom is different: its observed failures are
policy-level and are therefore handled by prompt instructions alone.
\section{Reproducibility and Artifact Details}
\label{app:reproducibility}
\subsection{Artifacts}
\label{app:artifacts}
Table~\ref{tab:artifact-summary} summarizes the external artifacts used in the
experiments. All artifacts are used for research evaluation. We do not release
or redistribute benchmark records, model outputs containing full trajectories,
or model weights. Our released code archive (footnote in
\S\ref{sec:budgets}) contains the code needed to reproduce the harness
optimization and analysis pipeline: the full PRISM implementation, all
baseline optimizer wrappers, benchmark runners, metric-aggregation scripts
including the \(\mathrm{RelLift}_{95}\) bootstrap, run configurations, and a
reproducibility guide with exact per-experiment commands.
\begin{table*}[t]
\centering
\small
\setlength{\tabcolsep}{4pt}
\renewcommand{\arraystretch}{1.08}
\caption{
External artifacts used in the experiments.
}
\label{tab:artifact-summary}
\begin{tabular}{
@{}
p{0.18\textwidth}
p{0.25\textwidth}
p{0.24\textwidth}
p{0.24\textwidth}
@{}
}
\toprule
Artifact
& Role in this paper
& Reported coverage
& License / terms handling \\
\midrule
BFCL multi-round
& Executable function-calling benchmark
& English tool-use tasks; 600 scorecard cases per split in our protocol
& Used through the public benchmark artifact and cited evaluation code; no benchmark redistribution. \\
\(\tau^2\)-Retail
& Simulated service-agent benchmark
& English retail workflows with dialogue state, policies, and tool-mediated order state; 74 scorecard instances per split
& Used under the repository license/terms for research evaluation; no benchmark redistribution. \\
\(\tau^2\)-Telecom
& Simulated service-agent benchmark
& English telecom workflows with dialogue state, policies, account state, and tool-mediated actions; 74 scorecard instances per split
& Used under the repository license/terms for research evaluation; no benchmark redistribution. \\
MIPROv2-style optimizer
& Prompt-only optimizer baseline
& Instruction/demo search over candidate harness prompts
& Cited as a baseline method; implementation details follow the optimizer description in Section~\ref{sec:optimizers}. \\
GEPA-style optimizer
& Prompt-only Pareto/reflection baseline
& Trace-reflective prompt proposal and Pareto retention
& Cited as a baseline method; implementation details follow the optimizer description in Section~\ref{sec:optimizers}. \\
Closed API LLMs
& Inner execution models and outer proposal models
& Model names reported in Sections~\ref{sec:experiments} and~\ref{sec:ablation}
& Accessed through provider APIs under their terms; no weights trained or redistributed. \\
\bottomrule
\end{tabular}
\end{table*}
\subsection{Compute, infrastructure}
\label{app:compute-setup}
All experiments are API-based. We do not train model weights or run local GPU
training. The main computational budget is therefore measured by task rollouts,
outer-model calls, token usage, and API cost rather than GPU hours. The main
experiments use \texttt{gpt-5.4-mini} as the fixed inner execution model and
\texttt{claude-opus-4.7} as the outer proposal model. Inner- and outer-model
ablations are reported in Tables~\ref{tab:inner-model-ablation-bfcl-retail}
and~\ref{tab:outer-model-ablation-bfcl-retail}.
For closed API models, provider-side serving infrastructure and model snapshots
are not fully observable. We therefore report the routed model names,
hyperparameters, experimental roles, software versions, and observed token/cost
budgets. The optimizer
search budget excludes final scorecard evaluation, as described in
Section~\ref{sec:budgets}. Token and cost accounting are reported in
Appendix~\ref{app:cost-methodology}.
Table~\ref{tab:implementation-parameters} lists the implementation and
evaluation parameters used in the main experiments.
\begin{table*}[t]
  \centering
  \small
  \setlength{\tabcolsep}{4pt}
  \renewcommand{\arraystretch}{1.08}
  \caption{
  Implementation and evaluation parameters used in the main experiments.
  }
  \label{tab:implementation-parameters}
  \begin{tabular}{
  @{}
  p{0.25\textwidth}
  p{0.65\textwidth}
  @{}
  }
  \toprule
  Parameter
  & Value \\
  \midrule
  Programming language
  & Python 3.12.10 \\
  Benchmark code versions
  & BFCL v4 (multi-turn); $\tau^2$-bench commit \texttt{337326e}
    (local install, version \texttt{0.2.1.dev0}) \\
  Experiment orchestration
  & \texttt{agent-harness-optimizer} v0.1.0 (this submission);
    \texttt{deepagents} v0.5.6;
    \texttt{litellm} v1.83.0 \\
  API clients
  & \texttt{openai} v2.34.0 (OpenAI / Azure);
    \texttt{anthropic} v0.98.1 (Anthropic / Bedrock);
    \texttt{langchain-anthropic} v1.4.3;
    \texttt{langchain-openai} v1.2.1 \\
  Randomness control
  & Four split seeds (0--3); up to four independent optimizer repeats per
    seed, with per-arm counts $n \le 16$ after the reliability screen
    (\S\ref{app:run-config}); bootstrap resampling
    ($N_\mathrm{boot}=5000$) for budgeted reliable lift \\
  Scorecard access
  & Final scorecard split held out from proposal, filtering, repair, and candidate
  selection \\
  Main inner model
  & \texttt{gpt-5.4-mini} \\
  Main outer model
  & \texttt{claude-opus-4.7} \\
  Main reported metrics
  & BFCL: $\mathrm{pass}^{1}$;
    $\tau^2$-Retail and $\tau^2$-Telecom: $\mathrm{pass}^{4}$ \\
  Failure handling
  & Crashes, invalid actions, protocol violations, repeated tool-call loops,
    and timeouts receive zero benchmark score \\
  Cost accounting
  & Logged pre-scorecard cost under Appendix~\ref{app:cost-methodology}; final held-out scorecard evaluation excluded \\
  \bottomrule
  \end{tabular}
  \end{table*}
\subsection{Baseline wrapper configurations}
\label{app:wrapper-config}
For reproducibility of the fairness mapping in \S\ref{sec:run-protocol}:
GEPA's published algorithm never consults data outside its training set
during search; here its training set is the repair split, it
receives the full repair split as its Pareto set and draws its feedback
minibatches from the repair split, and the gate split is consulted once, at
final acceptance. MIPROv2 follows the DSPy default configuration, with its
dedicated validation split taken as the last 20\% of the repair split; trials
are scored on repair-split minibatches, and the gate split is consulted once,
at final acceptance. BH maintains a single incumbent and consults the gate
split at every hill-climbing
iteration, accepting an edit iff gate pass rate improves. The GEPA-MW and
MIPROv2-MW variants receive the three middleware edit patterns of
\S\ref{sec:optimizers} exactly as PRISM's middleware mutation slot does,
under the identical protocol and budget. All wrapper configurations are
included in the released code archive.
\subsection{Model Parameters and Run Configuration}
  \label{app:run-config}
  \paragraph{Inner model (scored agent).}
  For BFCL, the inner model is called with \texttt{temperature=0.0} and
  \texttt{seed=42} (set explicitly in
  \texttt{agent\_harness\_optimizer/benchmarks/\_bfcl\_runner.py}).
  For $\tau^2$-bench, no temperature override is applied; the provider default
  (\texttt{temperature=1.0} for both GPT-5.4-mini via Azure and Claude models via
  Bedrock)
  is used, consistent with the $\tau^2$-bench evaluation protocol.
  The $\tau^2$-bench user simulator is called with \texttt{seed=42}.
  \paragraph{Outer model (proposer).}
  No temperature or \texttt{top\_p} is set explicitly in the codebase
  (\texttt{agent\_harness\_optimizer/utils/llm.py} calls \texttt{init\_chat\_model} with no sampling
  parameters). The provider default for Claude Opus 4.7 is \texttt{temperature=1.0}.
  All outer model calls are thus at provider default, which for Anthropic models
  corresponds to moderately high diversity (\texttt{temperature=1.0}, no
  \texttt{top\_p}
  override).
  \paragraph{Context and generation limits.}
  Per-rollout wall-clock timeouts are 5 minutes for BFCL
  (\texttt{wall\_time\_s=300}) and 10 minutes for
  the $\tau^2$ domains (\texttt{wall\_time\_s=600}).
  The outer model is capped at \texttt{--outer-max-turns 300} tool-use turns per
  optimization run. No \texttt{max\_tokens} cap is applied to the outer model;
  the inner model per-case token budget defaults to 500{,}000 tokens (BFCL) and
  100{,}000 tokens ($\tau^2$-bench).
  \paragraph{Parallelism.}
  BFCL inner evaluations use \texttt{asyncio.Semaphore(8)} (8 concurrent cases).
  $\tau^2$-bench runs cases via \texttt{asyncio.to\_thread}, also limited to 8
  concurrent simulations by \texttt{asyncio.Semaphore(8)}.
  \paragraph{Run exclusion.}
  Runs whose final repair-split reliability falls below 0.5 are excluded as
  infrastructure failures (API crashes, rate-limit storms) rather than
  optimizer outcomes; all per-arm statistics are computed over the remaining
  runs, so per-arm denominators can be smaller than the design maximum of 16.
  \paragraph{Optimizer-specific settings.}
  PRISM uses \texttt{generations=10}, \texttt{mutations\_per\_gen=3},
  \texttt{population\_cap=5}. BetterHarness uses \texttt{max\_iterations=10}.
  MIPROv2 uses \texttt{num\_candidates=10} with \texttt{num\_trials=34} and
  \texttt{minibatch\_size=25} on BFCL, and \texttt{num\_trials=5} with
  \texttt{minibatch\_size=4} on the $\tau^2$ domains.
  All optimizers share \texttt{--train-cases 100} / \texttt{--holdout-cases 100}
  for BFCL and \texttt{--train-cases 20} / \texttt{--holdout-cases 20} for
  $\tau^2$-bench benchmarks (each $\tau^2$ domain's 114-task base pool splits
  into 20 repair, 20 gate, and 74 scorecard cases). The component-ablation arms
  of \S\ref{sec:ablation-component} use these same settings.
\subsection{Uncertainty of \texorpdfstring{\(\mathrm{RelLift}_{95}\)}{RelLift95}: confidence intervals and subsampling}
\label{app:subsampling}
Table~\ref{tab:rellift-uncertainty} reports, for the \textsc{PRISM-MW} arm
on each benchmark, the percentile-bootstrap 95\% confidence interval on the
\(\mathrm{RelLift}_{95}\) estimate (outer bootstrap over the observed runs,
1{,}000 resamples) and a subsampling sensitivity analysis: the mean and
standard deviation of the estimate recomputed on random subsamples of
\(n \in \{16, 12, 8\}\) runs (200 subsamples per size, drawn without
replacement). A stable estimator moves little as runs are removed. The
\textsc{PRISM} estimates are stable under this check: subsample means match
the full-sample estimate at every size, the sd grows to at most
2.4 points at \(n{=}8\), and all three confidence intervals exclude
zero, so the positive-\(\mathrm{RelLift}_{95}\) claim does not rest on the
point estimates alone. Baseline
arms are reported as point estimates in
Table~\ref{tab:main-scorecard-results}; the same intervals can be computed
for any arm from its run tuples with \texttt{scripts/rellift.py} in the
released archive, which also produces all values here.
\begin{table*}[t]
\centering
\scriptsize
\setlength{\tabcolsep}{4.0pt}
\renewcommand{\arraystretch}{1.05}
\caption{
How much the \textsc{PRISM-MW} \(\mathrm{RelLift}_{95}\) estimates of
Table~\ref{tab:main-scorecard-results} can be trusted, checked two ways.
The 95\% CI column is a percentile-bootstrap confidence interval on each
estimate. The last three columns recompute the estimate on random
subsamples of \(n\) runs (mean \(\pm\) sd over 200 subsamples); a stable
estimator moves little as runs are removed. The full-pool column has sd
zero by construction, since there is only one full sample. Baseline arms
are reported as point estimates (\S\ref{sec:budgeted-estimator}); the
released estimator computes the same quantities for any arm.
}
\label{tab:rellift-uncertainty}
\begin{tabular}{@{} l l r c c c c @{}}
\toprule
Benchmark & Optimizer & \(\mathrm{RelLift}_{95}\) & 95\% CI
& \(n{=}16\) & \(n{=}12\) & \(n{=}8\) \\
\midrule
BFCL & PRISM-MW & 12.0 & [11.0, 15.8] & 12.0 \(\pm\) 0.0 & 12.0 \(\pm\) 0.1 & 12.0 \(\pm\) 2.2 \\
\(\tau^2\)-Retail & PRISM-MW & 10.1 & [5.4, 21.6] & 10.1 \(\pm\) 0.0 & 10.1 \(\pm\) 0.2 & 10.1 \(\pm\) 2.4 \\
\(\tau^2\)-Telecom & PRISM-MW & 7.8 & [6.8, 12.2] & 7.8 \(\pm\) 0.0 & 7.8 \(\pm\) 0.0 & 7.8 \(\pm\) 1.2 \\
\bottomrule
\end{tabular}
\end{table*}
\section{Token Tracking, Model Pricing, and Cost Methodology}
\label{app:cost-methodology}
This appendix defines the monetary quantities used in the
\emph{Cost/run (inner)} and \emph{Cost/run (all)} columns and in the
cost-based budgeted \(\mathrm{RelLift}_{95}(B)\) curves. These costs are reported only as
implementation-level diagnostics. They are not used to train, filter, or select
harnesses, and the final held-out scorecard evaluation is excluded.
\paragraph{Scope of the cost quantity.}
For each optimizer run \(i\), we record pre-scorecard model calls made during
search and price them under the rate-card convention below; final scorecard
evaluation is always excluded. Two scopes of this quantity appear in the
paper. The \emph{Cost/run (inner)} column of
Table~\ref{tab:main-scorecard-results} and the cost-based budgeted
\(\mathrm{RelLift}_{95}(B)\) curves sum \emph{inner-model task-rollout
calls only}, matching the controlled experimental budget, which is the
fixed inner-model rollout budget. The \emph{Cost/run (all)} column of the
model-role ablation tables
(Tables~\ref{tab:inner-model-ablation-bfcl-retail}
and~\ref{tab:outer-model-ablation-bfcl-retail}) sums all search-time model
calls, \emph{including} outer-proposer calls for harness modification,
which is why its figures under the main configuration exceed the main
table's inner-only figures (\$142 vs.\ \$103 per BFCL run). Monetary cost is a diagnostic in both scopes rather than the primary
fairness constraint.
\paragraph{Conservative rate-card convention.}
Let \(M_i\) be the set of model calls logged before scorecard evaluation in run
\(i\). For each model \(m\), let \(T^{(i,m)}_{\mathrm{in}}\) and
\(T^{(i,m)}_{\mathrm{out}}\) be the logged input and output token counts. We
compute
\[
C_i =
\sum_{m \in M_i}
\frac{
T^{(i,m)}_{\mathrm{in}} p^{(m)}_{\mathrm{in}}
+
T^{(i,m)}_{\mathrm{out}} p^{(m)}_{\mathrm{out}}
}{10^6}.
\]
The rates \(p^{(m)}_{\mathrm{in}}\) and \(p^{(m)}_{\mathrm{out}}\) are the
conservative rates used by our implementation-level cost calculation: when both
an internal route-card price and a public standard list price are available, we
use the larger value. We do not apply cached-token, batch, flex, reserved
capacity, or enterprise discounts. For models or routes without a public rate
card, we use the internal route-card rate and mark the source accordingly.
\paragraph{Interpretation.}
The resulting costs compare logged search overhead under our implementation.
They should not be interpreted as complete end-to-end deployment costs,
provider-independent prices, or a claim that all production inference costs are
captured. In particular, online serving costs, infrastructure costs, human
review costs, and final held-out scorecard evaluation are outside this cost
quantity.
\subsection{Model Rate Card Used for Logged Cost}
\label{app:model-rate-card}
Table~\ref{tab:model-rate-card} lists the token prices used in our logged cost
calculation. The table is a rate card for this paper's implementation-level
accounting, not a universal provider pricing catalog.
\begin{table*}[t]
\centering
\scriptsize
\setlength{\tabcolsep}{3.2pt}
\renewcommand{\arraystretch}{1.05}
\caption{
Token prices used for the logged cost columns, in USD per million input or
output tokens. Prices follow the conservative no-discount convention of
Appendix~\ref{app:cost-methodology}: no cached-token, batch, flex,
reserved-capacity, or enterprise discounts. Where a provider lists more
than one standard price (short vs.\ long context, or regional variation),
we take the higher one. Where an internal route-card rate exceeds the
public standard rate, the internal rate is used and marked as internal.
}
\label{tab:model-rate-card}
\resizebox{\textwidth}{!}{
\begin{tabular}{
@{}
p{0.23\textwidth}
p{0.18\textwidth}
p{0.13\textwidth}
r
r
p{0.18\textwidth}
p{0.14\textwidth}
@{}
}
\toprule
Model ID
& Common name
& Route
& Input
& Output
& Role
& Rate source \\
\midrule
\multicolumn{7}{@{}l}{\textit{Outer proposer models}} \\
\midrule
\texttt{claude-opus-4-7}
& Claude Opus 4.7
& Bedrock
& \$5.00
& \$25.00
& Outer, primary
& public standard \\
\texttt{claude-sonnet-4-6}
& Claude Sonnet 4.6
& Bedrock
& \$3.00
& \$15.00
& Outer, ablation
& public standard \\
\texttt{gpt-5.5}
& GPT-5.5
& Azure OpenAI
& \$10.00
& \$45.00
& Outer, ablation
& public standard max-context \\
\texttt{kimi-k2.5}
& Kimi K2.5
& Bedrock
& \$0.72
& \$3.60
& Outer, ablation
& public Bedrock max-region \\
\texttt{qwen3-32b-coder}
& Qwen3 Coder 30B A3B
& Bedrock
& \$0.1545
& \$0.6180
& Outer, ablation
& public Bedrock standard \\
\texttt{gpt-oss-120b}
& GPT-OSS 120B
& Bedrock
& \$0.1545
& \$0.6180
& Outer, ablation
& public Bedrock standard \\
\midrule
\multicolumn{7}{@{}l}{\textit{Inner execution models, used for rollout accounting and ablations}} \\
\midrule
\texttt{gpt-5.4-mini}
& GPT-5.4 Mini
& Azure OpenAI
& \$0.75
& \$4.50
& Inner, primary
& public standard \\
\texttt{gpt-5.4}
& GPT-5.4
& Azure OpenAI
& \$5.00
& \$22.50
& Inner, ablation
& public standard max-context \\
\texttt{gpt-5-nano}
& GPT-5 Nano
& Azure OpenAI
& \$0.05
& \$0.40
& Inner, ablation
& public standard \\
\texttt{qwen3-32b}
& Qwen3 32B
& Bedrock
& \$0.1545
& \$0.6180
& Inner, ablation
& public Bedrock standard \\
\texttt{gpt-oss-20b}
& GPT-OSS 20B
& Bedrock
& \$0.0721
& \$0.3090
& Inner, ablation
& public Bedrock standard \\
\bottomrule
\end{tabular}
}
\end{table*}
\end{document}